\documentclass[]{ailab}

\usepackage{amsmath,amssymb}
\usepackage{graphicx}
\graphicspath{{figures/}{.}}
\usepackage{tikz}
\usepackage{svg}
\usetikzlibrary{arrows.meta,positioning}
\DeclareRobustCommand{\hficon}{%
  \raisebox{-0.60ex}{%
    \includegraphics[height=1.32em]{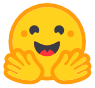}%
  }%
}

\DeclareRobustCommand{\githubicon}{%
  \raisebox{-0.20ex}{%
    \includegraphics[height=1.05em]{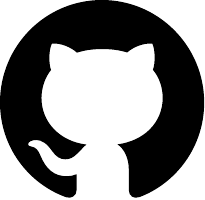}%
  }%
}
\usepackage{booktabs}
\usepackage{multirow}
\usepackage{rotating}
\usepackage{natbib}
\usepackage{xspace}
\usepackage{float}
\usepackage{xcolor}
\usepackage{ragged2e}
\usepackage{listings}
\usepackage{booktabs}
\usepackage{multirow}
\setcitestyle{square,comma,numbers,sort&compress}

\providecommand{\ours}{\textsc{RST}\xspace}

\title{\centering Recursive Synthesis for Long-Horizon Terminal Tasks}

\author[1,2]{Zhongzhi Li\textsuperscript{*}}
\author[1]{Yucheng Shi\textsuperscript{*}\textsuperscript{\(\dagger\)}}
\author[1,3]{Zongxia Li\textsuperscript{*}}
\author[1,4]{Ruhan Wang}
\author[5]{Anhao Li}
\author[2]{Jiaxi Li}
\author[1,6]{Junyao Yang}
\author[1]{Lei Ke}
\author[7]{Zixun Huang}
\author[8]{Yu Wang}
\author[2]{Jingyuan Huang}
\author[9]{Ninghao Liu}
\author[1]{Haitao Mi}
\author[1]{Leowei Liang}

\affiliation[1]{Tencent HY LLM Frontier}
\affiliation[2]{University of Georgia}
\affiliation[3]{University of Maryland, College Park}
\affiliation[4]{Indiana University Bloomington}
\affiliation[5]{University of Minnesota, Twin Cities}
\affiliation[6]{National University of Singapore}
\affiliation[7]{University of Pennsylvania}
\affiliation[8]{University of Illinois at Chicago}
\affiliation[9]{Hong Kong Polytechnic University}

\resource{%
  \hficon\quad
  \href{https://huggingface.co/collections/Zhongzhi1228/recursive-synthesis-for-long-horizon-terminal-tasks}
       {Hugging Face:  Recursive Synthetic Terminal Tasks}
  \hspace{1.8em}
  \githubicon\quad
  \href{https://zhongzhi660.github.io/recursive-verified-synthesis-site/?case=jobs-diff-01-3341b098}
       {Project Website}
}

\email{zl22754@uga.edu, zhongzhili@global.tencent.com}

\contribution{\textsuperscript{*}Equal contribution}
\contribution{\textsuperscript{\(\dagger\)} Project lead.}

\abstract{%
\justifying
\setlength{\parfillskip}{0pt}

High-quality long-horizon training data for terminal agents is costly and difficult to scale, as manual authoring is expensive and direct LLM generation often produces inconsistent instructions, environments, solutions, and verifiers.
We present Recursive Synthetic Terminal Tasks (\ours{}), a recursive verified synthesis framework for constructing long-horizon terminal-agent tasks at scale.
Starting from verified seed tasks, \ours{} extends the reference solution, realigns the verifier and instruction to the new workflow, validates the result in a fresh sandbox, and reuses accepted tasks as seeds for subsequent rounds.
Across fifteen recursive rounds, \ours{} produces 37,484 synthesized terminal-agent tasks at roughly \$0.05 per task.
Task difficulty increases substantially over rounds: the median reference solution grows from 67 to 374 lines, the median number of executed commands grows from 40 to 244, and DeepSeek-V4-Pro pass@4 drops from 90\% at $R_1$ to 2.5\% at $R_{15}$.
To demonstrate training utility, we collect rejection-sampled Qwen3.5 trajectories on the synthesized tasks and use them for supervised fine-tuning.
Fine-tuning on these trajectories improves Qwen3.5-27B and Qwen3.5-122B-A10B by up to 10\% on Terminal-Bench~2, Terminal-Bench Hard, and Long-Horizon Terminal Bench, while agentic PPO lifts Qwen3.5-27B to 49.44\%, 32.00\%, and 22.07\% on the three benchmarks, 
corresponding to relative gains of 20.0\%, 41.2\%, and 21.9\% over the base model.
Moreover, after 15 rounds, the recursion shows no ceiling: synthesis yield and validation rates remain stable as difficulty keeps climbing, indicating that the process can continue well beyond the scale reported here.
\par
\vspace{0.2em}
{\centering\begin{minipage}{0.98\linewidth}
  \centering
  \captionsetup{font=footnotesize,skip=2pt}
  \includegraphics[width=\linewidth]{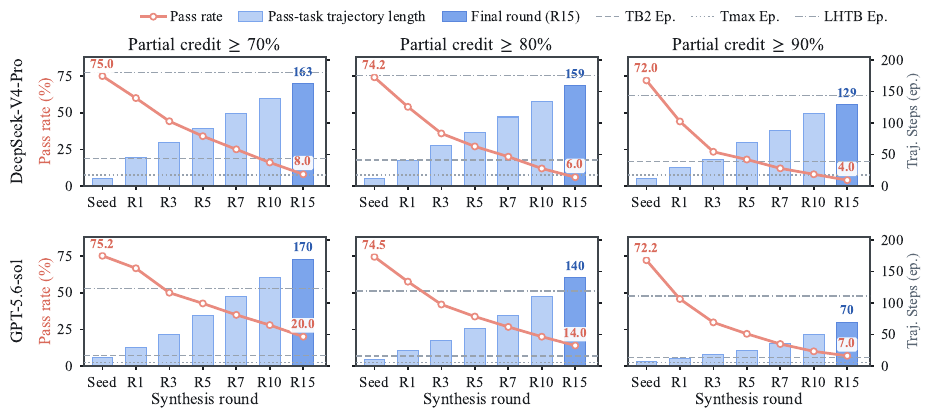}
  \captionsetup[figure]{position=bottom,skip=4.5pt}
  \captionof{figure}{Pass rate and trajectory length across recursive synthesis rounds for DeepSeek-V4-Pro and GPT-5.6-sol under 70\%, 80\%, and 90\% partial credit thresholds.}
  \label{fig:fixed-model-overview}
\end{minipage}\par}
\par
}

\begin{document}
\thispagestyle{firstheader}
\maketitle

\section{Introduction}
\label{sec:intro}

Large language models (LLMs) are increasingly used as agents rather than one-shot generators~\citep{yang2026stale,li2026less}.  
Instead of producing a single answer, they are expected to execute complex workflows and work toward a goal~\citep{wang2026harness}.
This shift from simple chatting to multi-step tasks is reflected in recent work on tool use and software agents~\citep{du2026survey,chowa2026language,ICLR2024_edac78c3} such as interactive benchmarks for web, computer-use, and long-horizon problem solving~\citep{liu2025agentbenchevaluatingllmsagents,mialon2023gaiabenchmarkgeneralai}. 
In terminal-based tasks, agents must coordinate repository context, command-line interaction, and execution feedback over extended workflows~\citep{yang2024sweagentagentcomputerinterfacesenable,merrill2026terminalbenchbenchmarkingagentshard,swemarathon_2026}.  
As a result, progress depends not only on stronger models, but also on harness~\cite{Harbor_Framework} and executable training data that teaches reliable long-horizon behavior~\cite{li2026longhorizonterminalbenchtestinglimitsagents}.

However, high quality long-horizon tasks are hard to obtain at scale~\cite{vidgen2026apexagents}. 
Useful training data for terminal agents must be verified through a runnable workspace with an instruction, meaningful execution feedback, and a reliable verifier to check success~\citep{ICLR2024_edac78c3,merrill2026terminalbenchbenchmarkingagentshard,Harbor_Framework}.  
Human-written tasks and successful trajectories are expensive to produce, often ranging from hundreds to thousands of dollars per task~\citep{yang2024sweagentagentcomputerinterfacesenable,wang2025openhandsopenplatformai,swemarathon_2026}.  
%
%
Existing execution-based benchmarks and recent benchmark builders suggest that parts of data sourcing and environment construction can be automated~\citep{ICLR2024_edac78c3,zan2025multiswebenchmultilingualbenchmarkissue,wang2025swebenchframeworkscalablegeneration}, but they do not yet provide a scalable way to build \textit{high quality} verified terminal-agent data.

In this paper, we propose \ours{}, a recursive framework for synthesizing verifiable terminal-agent tasks.  
\ours{} starts from an existing seed task, then extends its reference solution to create a longer and more demanding solution.  
It then updates the verifier and instruction to match the new solution and validates the complete task in a sandbox. 
Valid tasks are used to collect training trajectories and become seeds for the next synthesis round. 
We use DeepSeek-V4-Pro~\cite{deepseekai2026deepseekv4} to recursively produce increasingly difficult tasks and corresponding trajectories for agent training.
%
To validate the usefulness of synthesized data, we conduct 15 rounds of recursive synthesis and produce 37,484 verified terminal tasks from 639 seed tasks.  
Later rounds require substantially more work: median solution length increases by 5.6$\times$ and command use by 6.1$\times$, while instruction length increases by only 1.4$\times$.  
When the same solvers are evaluated across rounds, pass rates under the strictest criterion fall from $72\%$ to 4\% for DeepSeek-V4-Pro and $72.2\%$ to 7\% for GPT-5.6-sol at Seed to $R_{15}$.  
Trajectories collected by self-rolling out Qwen3.5 on these tasks improve Qwen3.5-27B and Qwen3.5-122B-A10B through supervised fine-tuning on Terminal-Bench~2, Terminal-Bench~Hard, and Long-Horizon Terminal Bench. 
We also show that PPO training further increases Qwen3.5-27B by $8.24\%$, $9.33\%$, and $3.97\%$ on the three benchmarks, respectively.

Our work shows the following insights:
\begin{enumerate}
  \item \textbf{Scalable, low-cost synthesis of long-horizon tasks.} Our recursive, solution-first synthesis framework in which every accepted task carries an executable proof of solvability.  From 639 seed tasks, fifteen rounds produce 37,484 verified terminal tasks at approximately \$0.05 per passed task, with no human authoring in the loop.  Median solution length grows 5.6$\times$ and command use 6.1$\times$, and successful GPT-5.6-sol trajectories on late-round tasks exceed 100 steps.
  \item \textbf{Consistent downstream improvement under standard training.}  Self-collected trajectories by Qwen3.5 improve Qwen3.5-27B and Qwen3.5-122B-A10B on Terminal-Bench~2, Terminal-Bench~Hard, and Long-Horizon Terminal Bench through plain supervised fine-tuning, with gains of up to 10 points. 
  \item \textbf{No observed ceiling.}  After fifteen recursive rounds, passed-task yield and candidate pass rates remain stable, per-round structural growth stays positive, and domain, operator, and rewrite-family diversity are all preserved, even as solver pass rates fall from above 70\% to 4--7\%.  The recursion shows no sign of saturation or collapse, and the pipeline depends on no particular seed domain or generation model, indicating that synthesis can continue to scale in both data quantity and quality.
\end{enumerate}


\section{Related Work}
\label{sec:related}

\paragraph{Terminal-agent benchmarks and long-horizon evaluation.}
Repository-level benchmarks such as SWE-Bench and Multi-SWE-Bench evaluate whether agents can modify codebases and satisfy executable tests~\citep{ICLR2024_edac78c3,zan2025multiswebenchmultilingualbenchmarkissue}.
SWE-Bench++ and SWE-Marathon extend this setting to broader and more demanding software-engineering tasks~\citep{wang2025swebenchframeworkscalablegeneration,swemarathon_2026}.
Terminal-Bench evaluates interactive command-line workflows, while Harbor provides a common interface for sandboxed execution and verifier-based grading~\citep{merrill2026terminalbenchbenchmarkingagentshard,Harbor_Framework}.
Long-Horizon-Terminal-Bench extends this setting to stateful terminal tasks that require hundreds of interactions and uses dense verifier-based rewards to measure partial progress~\citep{li2026longhorizonterminalbenchtestinglimitsagents}.
TerminalWorld constructs validated terminal tasks from real interaction records, including workflows that require more than fifty steps~\citep{chu2026terminalworld}.
Broader suites evaluate agents across web, desktop, and multi-application settings~\citep{liu2025agentbenchevaluatingllmsagents,mialon2023gaiabenchmarkgeneralai,zhou2024webarenarealisticwebenvironment}.
Recent studies further emphasize persistent state, delayed feedback, and repeated tool use in long-horizon evaluation~\citep{xie2024osworldbenchmarkingmultimodalagents,kwa2026measuringaiabilitycomplete}.
Agents are commonly evaluated through shared scaffolds such as SWE-agent and OpenHands~\citep{yang2024sweagentagentcomputerinterfacesenable,wang2025openhandsopenplatformai}.
These works primarily define evaluation settings, whereas \ours{} recursively expands executable tasks and uses them to collect training trajectories.

\paragraph{Synthetic tasks and verifiable environments.}
Recent work generates tasks and environments for tool use, coding, and agent training~\citep{NEURIPS2025_a5a305fa,xia2025agent0unleashingselfevolvingagents,zhu2025sweplayground,wu2026large,ren2026self,kulikov2026autodata}.
Other systems reduce human authoring by scaling web, desktop, and general digital environments~\citep{song2026envscaler,fang2025webevolver,wang2025llmsscalablegeneralpurposesimulators}, while continual world generation and automatic environment construction extend this direction to evolving interaction settings~\citep{zhang2026infiniteweb,wu2026autowebworld,rlve}.
For terminal agents, Endless Terminals and LiteCoder-Terminal construct executable training environments~\citep{gandhi2026endlessterminalsscalingrl,peng2026litecoderterminalscalinglonghorizonterminal}, and CLI-Universe and SETA study their use in supervised and reinforcement learning~\citep{hua2026cliuniverseverifiabletasksynthesis,shen2026setascalingenvironmentsterminal}.
Among the compared methods, \ours{} is the only one that repeatedly uses validated task bundles as seeds across multiple synthesis generations.



\paragraph{Recursive synthesis and agent self-improvement.}
Self-training and self-play improve reasoning models by generating new examples and filtering them with feedback~\citep{zelikman2022star,li2026mm,he2025visplay,singh2024humandatascalingselftraining,chen2024self, li2026online}.
Self-rewarding and verifier-guided methods extend this approach through learned or executable reward signals~\citep{yuan2025selfrewardinglanguagemodels,li2025self,hosseini2024v,li2026mits,huang2026rzeroselfevolvingreasoningllm}.
However, recursive training can become unstable when filtering is weak or the reward is misspecified~\citep{zhao2025absolutezeroreinforcedselfplay,prasad2024self,fu2025neurips-selfverification}.
Several recent methods bring task evolution to agent settings.
SETA adapts the difficulty and diversity of terminal environments~\citep{shen2026setascalingenvironmentsterminal}.
BenchEvolver modifies executable solutions before deriving harder coding tasks and tests~\citep{wu2026benchevolverfrontiertasksynthesis}.
TRACE evolves agent tasks through validated and reproducible trajectories~\citep{guo2026selfevolvingbenchmarkssynthesizingagent}.
\ours{} recursively evolves complete terminal tasks, including the workspace, reference solution, verifier, and public instruction.
After each rewrite, these components are realigned and validated together.
Accepted tasks are incorporated into subsequent synthesis seed pools and directly constitute the training task pool for verifier-based reinforcement learning.  Successful rollouts collected on these tasks are retained as supervised fine-tuning trajectories.

\begin{table*}[t]
\centering
\caption{Comparison of representative task datasets and environment-generation pipelines. Task and trajectory counts are reported separately because executable environments and agent interaction traces are distinct data units. Prior-work properties follow the corresponding papers and releases.}
\label{tab:dataset-comparison}
\small
\newcommand{\compyes}{\textcolor{green!50!black}{\ensuremath{\checkmark}}}
\newcommand{\compno}{\textcolor{red!70!black}{\ensuremath{\times}}}
\newcommand{\compmix}{\textcolor{orange!80!black}{\ensuremath{\triangle}}}
\setlength{\tabcolsep}{3.6pt}
\renewcommand{\arraystretch}{1.32}

\resizebox{\textwidth}{!}{%
\begin{tabular}{llrrccccc}
\toprule
Dataset
& Domain
& \shortstack{Tasks\textsuperscript{\(\dagger\)}}
& \shortstack{Trajectories\textsuperscript{\(\ddagger\)}}
& Grounded?
& Executable?
& \shortstack{Adaptive?}
& RL-Validated?
& \shortstack{Reseeding?} \\
\midrule

WizardLM~\citep{ICLR2024_82eec786}
& General
& --
& 250,000\textsuperscript{a}
& \compno & \compno & \compno
& \compno~(SFT) & \compno \\

SWE-Gym~\citep{pan2025trainingsoftwareengineeringagents}
& SWE
& 2,438
& N/R
& \compyes & \compyes & \compno
& \compno~(SFT) & \compno \\

OpenThoughts-Agent-v1-SFT~\citep{raoof2026openthoughtsagentdatarecipesagentic}
& Terminal
& N/R
& 15,209
& \compmix & \compyes & \compno
& \compno~(SFT) & \compno \\

TermiGen~\citep{zhu2026termigenhighfidelityenvironmentrobust}
& Terminal
& 3,500+
& 3,291
& \compno & \compyes & \compno
& \compno~(SFT) & \compno \\

RLVE-Gym~\citep{rlve}
& Math/Algo
& 400
& N/R
& \compyes & \compno & \compyes
& \compyes~(DAPO) & \compno \\

ScaleEnv~\citep{tu2026scaleenvscalingenvironmentsynthesis}
& Tool-Use
& 2,560
& N/R
& \compno & \compyes & \compno
& \compyes~(GRPO) & \compno \\

Endless Terminals~\citep{gandhi2026endlessterminalsscalingrl}
& Terminal
& 3,255
& N/R
& \compno & \compyes & \compno
& \compyes~(PPO) & \compno \\

Terminal-Corpus~\citep{pi2026dataengineeringscalingllm}
& Terminal
& N/R
& 254,000+
& \compyes & \compyes & \compno
& \compno~(SFT) & \compno \\

TMax-15K~\citep{ivison2026tmaxsimplerecipeterminal}
& Terminal
& 14,600
& N/R\textsuperscript{b}
& \compno & \compyes & \compno
& \compyes~(GRPO) & \compno \\

SETA~\citep{shen2026setascalingenvironmentsterminal}
& Terminal
& 4,567
& 1,112\textsuperscript{c}
& \compyes & \compyes & \compyes
& \compyes~(GRPO) & \compno \\

\midrule
\textbf{\ours{} (ours)}
& \textbf{Terminal}
& \textbf{37,484}
& \textbf{327,189}
& \compyes & \compyes & \compyes
& \compyes~\textbf{(SFT, PPO)} & \compyes \\

\bottomrule
\end{tabular}%
}

\vspace{0.3em}
\begin{minipage}{\textwidth}
\footnotesize
\textsuperscript{\(\dagger\)}Tasks. counts reusable task or environment instances rather than model executions.
\textsuperscript{\(\ddagger\)}Trajectory counts collected agent interaction traces. N/R indicates that no fixed trajectory-corpus size is reported.
\textsuperscript{a}WizardLM reports single-turn instruction--response examples rather than terminal-agent trajectories.
\textsuperscript{b}TMax separately reports 16.5k SFT trajectories generated from an additional 2.2k-environment warm-start set. These are not counted as trajectories from TMax-15K.
\textsuperscript{c}SETA retains 1,112 successful SFT trajectories from 1,488 collected rollouts.
\compmix~indicates a mixture of grounded and synthetically generated sources.
\end{minipage}
\end{table*}

\section{Preliminaries}
\label{sec:prelim}

We formalize the executable terminal-task representation, the rollout protocol, and the criteria used to accept synthesized tasks.
A terminal task is a self-contained executable problem.  It contains the following components:
\begin{itemize}
  \item \texttt{instruction.md}: the public task description;
  \item \texttt{task.toml}: runtime metadata and configuration;
  \item \texttt{environment/Dockerfile}: the initial environment and workspace;
  \item \texttt{solution/solve.sh}: the reference solution;
  \item \texttt{tests/test.sh} and \texttt{tests/test\_state.py}: the private verifier.
\end{itemize}

Model interaction and grading are managed through Harbor using Terminus-2 as the agent harness~\citep{Harbor_Framework}.  For each rollout, Harbor creates an isolated sandbox and provides Terminus-2 with the public instruction and initialized workspace.  Terminus-2 may inspect files, execute commands, and modify the workspace, but it cannot access the reference solution or private verifier.  After the interaction ends, Harbor runs the verifier on the final workspace state.  The verifier checks task outcomes rather than a fixed command sequence, allowing different valid solutions to receive credit.

A synthesized task is accepted only when it satisfies two conditions.  First, the reference solution must pass the private verifier in a fresh sandbox; we refer to this as \emph{oracle validity}.  Second, every requirement checked by the verifier must be stated in the public instruction or inferable from the workspace; we refer to this as \emph{contract validity}.  The first condition establishes that the task is executable, while the second prevents private tests from introducing requirements hidden from the agent.

We denote the fifteen synthesis rounds by $R_1,\ldots,R_{15}$.  The process begins with 639 verified bootstrap tasks, and the bootstrap tasks themselves are not counted as a synthesis round.  For each $r=2,\ldots,15$, selected tasks from $R_{r-1}$ are transformed and validated, and the accepted candidates form $R_r$.  Accepted tasks may seed subsequent synthesis rounds and directly constitute the task pool for verifier-based reinforcement learning.  Successful rollouts collected from these tasks are retained as supervised fine-tuning trajectories.
\section{Method}
\label{sec:method}

\ours{} constructs terminal-agent training tasks over multiple synthesis rounds.  
In each round, accepted tasks from the preceding round are selected as seeds.  For each seed, the reference solution is extended, the verifier and public instruction are updated to match the new workflow, and the resulting task is validated in a fresh sandbox.  This executable task format follows repository-level benchmarks~\citep{ICLR2024_edac78c3,zan2025multiswebenchmultilingualbenchmarkissue,wang2025swebenchframeworkscalablegeneration} and terminal benchmarks~\citep{merrill2026terminalbenchbenchmarkingagentshard}.  Accepted tasks enter both the next synthesis seed pool and the reinforcement-learning task pool, while successful rollouts provide supervised fine-tuning trajectories.

\begin{figure}[t]
    \centering
    \includegraphics[width=\linewidth]{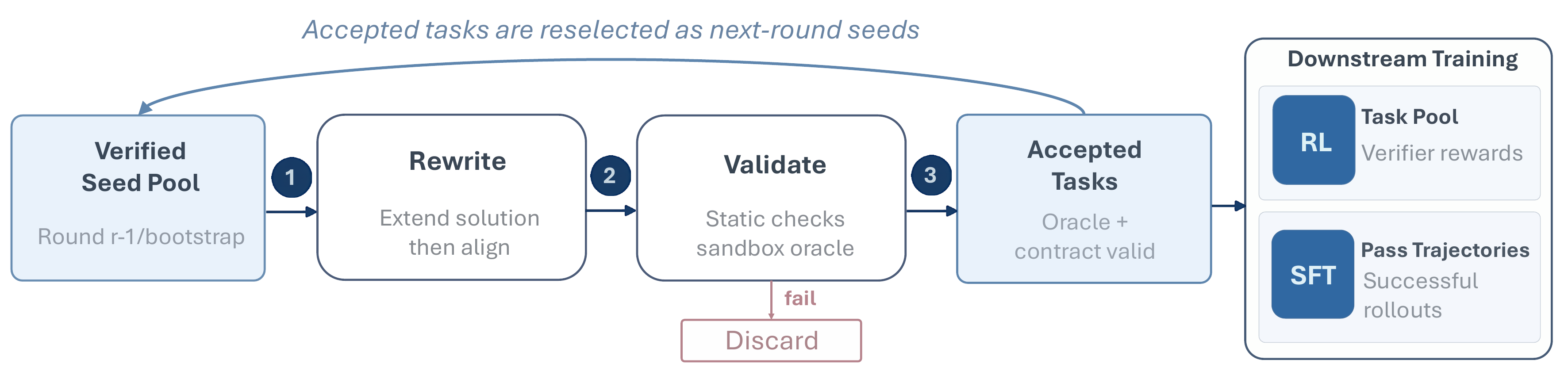}
    \caption{Recursive task synthesis and agent training in \ours{}. Accepted tasks seed subsequent synthesis rounds and form the task pool for reinforcement learning, while successful rollouts provide supervised fine-tuning trajectories.}
    \label{fig:coevolution}
\end{figure}

Figure~\ref{fig:coevolution} summarizes the interaction between task synthesis and model training.  Starting from a verified seed, the framework extends the reference solution and updates the verifier and public instruction to construct a harder candidate.  The candidate is validated in a sandbox; unsolvable tasks are discarded, while accepted tasks enter the verified pool.  This pool supplies seeds for subsequent synthesis rounds and tasks for verifier-based RL, while successful rollouts provide SFT trajectories.  

Each synthesis round follows four stages.  First, \ours{} selects a feasible rewrite operator and defines the expected outcomes based on the seed task. It then extends the executable workflow and updates the solution, verifier, instruction, and environment to keep them consistent. Each candidate undergoes static checks, anti-shortcut and leakage audits, and validation in a fresh sandbox; recoverable failures receive bounded repair. 
Finally, accepted tasks are selected and cohort caps to form the next-round seed pool.\footnote{Prompt templates and implementation details are provided in Appendix~\ref{app:synth-implementation} and Appendix~\ref{app:engineering}.}

\begin{figure}[t]
    \centering
    \includegraphics[width=\linewidth]{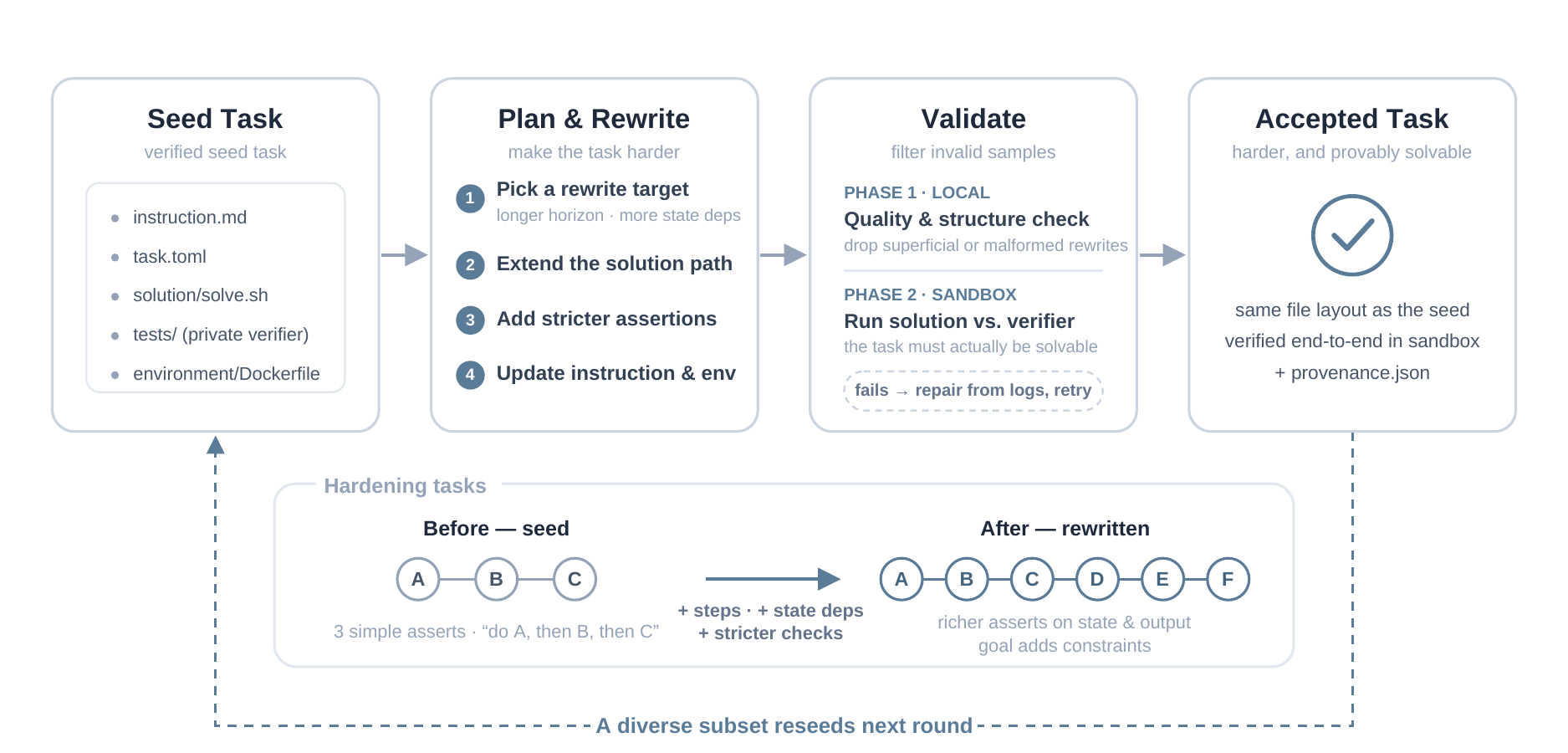}
    \caption{Detailed pipeline of one recursive synthesis round. A verified seed task undergoes target selection, staged rewriting, local and sandbox validation, and diversity-controlled reseeding.}
    \label{fig:pipeline}
\end{figure}

\subsection{Seed Pool and Diversity-Capped Selection}
\label{sec:method-seeds}

\begin{figure}[t]
  \centering
  \includegraphics[width=\linewidth]{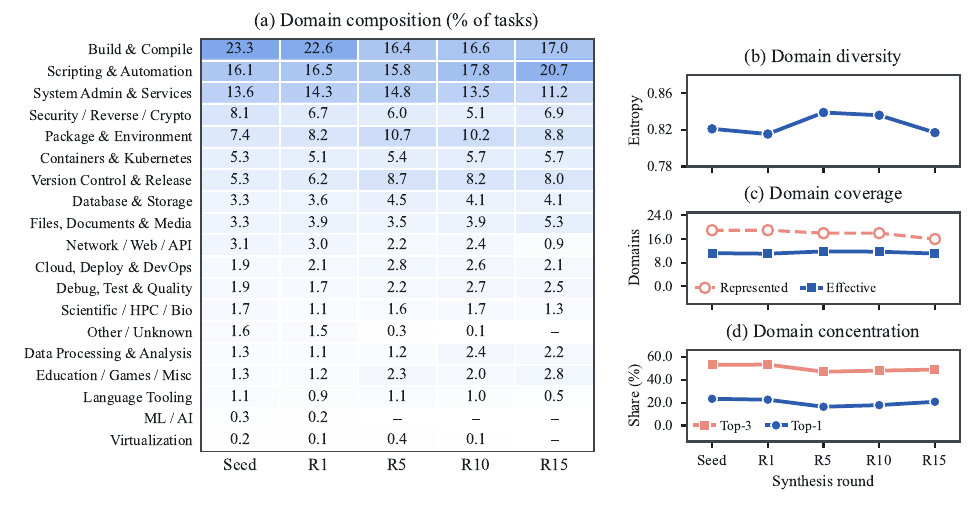}
  \caption{Domain composition and stability from the 639 bootstrap seeds through recursive synthesis. {(a)} Ancestral-domain proportions for the Seed pool and oracle-passed tasks at $R_1$, $R_5$, $R_{10}$, and $R_{15}$. {(b)} Distribution evenness measured by normalized Shannon entropy. {(c)} Domain breadth measured by the represented and effective numbers of domains. {(d)} Domain concentration measured by the proportions of tasks belonging to the Top-1 and Top-3 domains. The stable entropy and effective domain count indicate that domain diversity is preserved through $R_{15}$, with no broad domain collapse.}
  \label{fig:seed-domain-dist}
\end{figure}

We initialize the synthesis pipeline with 639 verified tasks as bootstrap seeds, sampled from TerminalWorld~\citep{chu2026terminalworld}, a dataset of validated terminal tasks constructed from real interaction records.  Applying the synthesis and validation process to these seeds produces 2,820 accepted tasks, which define $R_1$; the 639 bootstrap seeds are not counted as a synthesis round.  For diversity analysis, the original category labels of the bootstrap seeds are consolidated into 19 domains.  As shown in Figure~\ref{fig:seed-domain-dist}, domain composition remains broadly distributed from the Seed pool through $R_{15}$: the largest domain remains below one quarter of the pool, normalized entropy changes only from 0.821 to 0.817, and the effective number of domains remains nearly unchanged (11.22 versus 11.09). These results show no evidence of broad domain collapse.\footnote{The complete domain mapping is provided in Appendix~\ref{app:seed-diversity}.}

For each subsequent round $R_r$, where $r=2,\ldots,15$, seeds are selected from the accepted pool of $R_{r-1}$.  Tasks missing required components are removed before selection.  The remaining candidates are selected under caps on parent lineage, category, rewrite family, and generation cohort.  These constraints prevent a small number of parents or rewrite patterns from dominating later rounds.  The selected tasks enter the next synthesis round, and the accepted children form $R_r$.

\subsection{Target Selection and Task Contract}
\label{sec:method-plan}

\begin{figure*}[tbp]
  \centering
  \begin{minipage}[t]{0.42\textwidth}
    \centering
    \includegraphics[width=\linewidth]{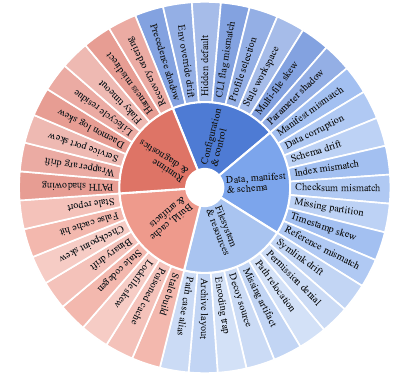}
    \caption{Conceptual taxonomy of terminal-agent primitives used to characterize capability coverage.  Ring labels are shortened; the implementation-level rewrite families and their 40 operators are defined separately in Appendix~\ref{app:operators}.}
    \label{fig:operator-taxonomy}
  \end{minipage}\hfill
  \begin{minipage}[t]{0.555\textwidth}
    \centering
    \includegraphics[width=\linewidth]{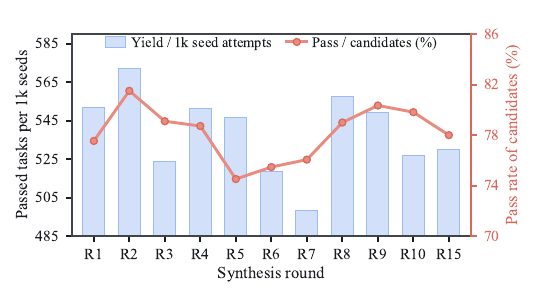}
    \caption{Passed-task yield per 1,000 seed attempts and candidate pass rate across recursive synthesis rounds. Both measures remain stable through $R_{15}$, with no systematic decline in synthesis throughput or validation success.}
    \label{fig:normalized-yield}
  \end{minipage}
\end{figure*}

Each synthesis round starts by selecting a suitable rewrite operator.  The pipeline inspects the seed task, including its files, tools, dependencies, and existing workflow, to determine which extensions are feasible.  It then selects one of the 40 operators shown in Figure~\ref{fig:operator-taxonomy}.  These operators are grouped into five families: Configuration and Control State; Data, Manifest, and Schema State; Filesystem and Resource Binding; Build, Cache, and Artifact State; and Runtime, Tooling, and Diagnostics.\footnote{Detailed operator definitions are provided in Appendix~\ref{app:operators}.}

Before modifying the task, the generator records a rewrite plan.  The plan defines the new required behavior, the corresponding changes to the reference solution, the intermediate and final outcomes checked by the verifier, and the information available to the agent through the instruction or workspace.  It also identifies shortcuts that the verifier must reject.  A plan is rejected if it introduces only cosmetic changes, adds checks unrelated to the executable workflow, or places required information only in private tests.  An approved plan guides the subsequent updates to the solution, verifier, instruction, and environment.

\subsection{Rewrite: Grow Solution, then Align}
\label{sec:method-rewrite}

The rewrite proceeds from executable behavior and runtime conditions to the public task specification. First, the generator extends \texttt{solve.sh} with additional operations, such as inspecting files, deriving values, invoking tools, managing state, producing intermediate artifacts, or validating final outputs. The environment is then modified to support these operations. Such modifications may install new dependencies and CLI tools, provide additional files, fixtures, or configuration, initialize services and processes, or adjust permissions, paths, resource limits, and runtime settings. Once the solution and environment define a complete execution path, the verifier is updated to check the resulting artifacts and state transitions while rejecting placeholders, hard-coded outputs, and omitted intermediate work. 
The public instruction is revised to state the new objective and identify required information that is not discoverable from the environment. Task metadata is updated when the modified environment requires different resources, timeouts, or execution settings.


\subsection{Validate: Local Filters and Sandbox Oracle}
\label{sec:method-validate}

Validation proceeds in two stages.  First, static checks reject candidates that are near-duplicates of their seeds, omit required files, contain invalid metadata, or expose private verifier details in the public instruction.  This stage removes obvious failures before allocating a sandbox.

Candidates that pass the static checks are evaluated in a fresh sandbox.  The environment is built from its initial state, the reference solution is executed, and the private verifier is run on the resulting workspace. When a failure is repairable, the validation logs are used for a limited number of repairs restricted to approved files.  
The candidate is then validated again; persistent failures are discarded.
We use \emph{oracle-passed} for candidates whose reference solution passes the private verifier in the sandbox.  A candidate is \emph{accepted} only after it also passes the instruction-verifier consistency checks.\footnote{The complete filters and repair procedure are provided in Appendix~\ref{app:filters-repair}.}

\section{Experiments}
\label{sec:experiments}


We evaluate \ours{} from two perspectives: the performance of the synthesis pipeline and the utility of the synthetic data for model training. The synthesis evaluation covers efficiency, structural growth, validity, diversity, and task difficulty across fifteen rounds. Efficiency is measured by normalized yield and generation cost; structural growth by changes in the solution and verifier; validity by instruction-verifier consistency; and diversity by rewrite-family balance, operator coverage, lineage retention, novelty, and near-duplicate similarity.  Task difficulty is evaluated on subsets from each round using DeepSeek-V4-Pro pass@4 and partial credit. Pass@4 measures full completion within four attempts, while partial credit is the fraction of verifier checks passed after a rollout. Training utility is evaluated through supervised fine-tuning of Qwen3.5-27B and Qwen3.5-122B-A10B and verifier-based reinforcement learning of Qwen3.5-27B. Terminal-Bench~2 evaluates broad terminal execution in standardized sandboxed environments with executable grading~\citep{merrill2026terminalbenchbenchmarkingagentshard}. We evaluate transfer to an independently constructed task distribution using \emph{Terminal-Bench Hard}, a 100-task subset drawn from TMax-15K~\citep{ivison2026tmaxsimplerecipeterminal}.  Long-Horizon Terminal Bench evaluates persistent, multi-stage terminal workflows with many dependent interactions and dense partial-credit grading~\citep{li2026longhorizonterminalbenchtestinglimitsagents}. To verify that the observed transfer is not explained by benchmark leakage, we conduct a task-description audit comparing samples from $R_1$, $R_5$, $R_{10}$, and $R_{15}$ with all 89 TB2 tasks, 100 Terminal-Bench Hard tasks, and 46 LHTB tasks (Table~\ref{tab:round_benchmark_overlap}). Under the normalized 13-token sliding-window criterion, no benchmark task matches any sampled synthesis round, and the maximum pairwise 5-gram Jaccard similarity remains below 0.009. Moreover, unigram Jensen--Shannon divergence increases from $R_1$ to $R_{15}$ for all three benchmarks, indicating that recursive synthesis produces an increasingly distinct task distribution rather than converging toward the benchmark distribution.

\begin{table*}[t]
\centering
\small
\setlength{\tabcolsep}{7pt}
\renewcommand{\arraystretch}{1.30}

\caption{Task-description distance and contamination analysis across recursive synthesis rounds. Unigram JSD measures the lexical-distribution difference between each round and the corresponding benchmark, while Max $J_5$ reports the largest pairwise 5-gram Jaccard similarity. Exact 13-token overlap reports the number of benchmark tasks containing at least one matching normalized 13-token window.}
\label{tab:round_benchmark_overlap}

\begin{tabular}{
  >{\centering\arraybackslash}m{1.4cm}
  >{\centering\arraybackslash}m{2.3cm}
  >{\centering\arraybackslash}m{1.5cm}
  >{\centering\arraybackslash}m{1.5cm}
  >{\centering\arraybackslash}m{1.5cm}
  >{\centering\arraybackslash}m{2.0cm}
}
\toprule
&
&
\multicolumn{3}{c}{Unigram JSD}
&
\\
\cmidrule(lr){3-5}
Round
& Median tokens
& TB2
& LHTB
& TB Hard
& Max $J_5$ \\
\midrule
$R_1$    & 87.5  & 0.358 & 0.441 & 0.331 & 0.0081 \\
$R_5$    & 104.0 & 0.387 & 0.461 & 0.359 & 0.0042 \\
$R_{10}$ & 116.0 & 0.426 & 0.484 & 0.394 & 0.0028 \\
$R_{15}$ & 121.5 & 0.433 & 0.485 & 0.406 & 0.0051 \\
\midrule
\multicolumn{6}{l}{
  Exact 13-token overlap:
  \quad TB2 $0/89$
  \quad LHTB $0/46$
  \quad TB Hard $0/100$
} \\
\bottomrule
\end{tabular}
\end{table*}

The synthesis process begins with 639 verified bootstrap tasks.  Applying the synthesis and validation pipeline to these tasks produces 2,820 accepted tasks, which define $R_1$; the bootstrap tasks are not counted as a synthesis round.  For each $r=2,\ldots,15$, a subset of the accepted tasks in $R_{r-1}$ is selected as the seed pool, and the accepted outputs of the new synthesis round form $R_r$.  The resulting $R_1$--$R_{15}$ pools contain 37,484 tasks.  Each submitted seed counts as one synthesis attempt, and synthesis yield is normalized per 1,000 attempts to support comparison across rounds.  Structural, validity, and diversity statistics are computed from the accepted task pools, while solver difficulty is evaluated on matched subsets.  All values are derived from the original generation and validation records.

\subsection{Design Ablation: Solution-First Generation and Recursive Seed Continuation}
\label{sec:exp-2x2-ablation}

Before evaluating the full recursive synthesis pipeline, we verify that the two central design choices of \ours{}, solution-first generation and recursive seed continuation, are individually and jointly beneficial.  We run a $2\times 2$ ablation over the two design axes.  The generation-order axis compares \emph{instruction-first} generation, where the public instruction is generated before the reference solution and verifier, against \emph{solution-first} generation, where the reference solution and environment are constructed first and the verifier and instruction are then aligned to the executed behavior.  The seed-update axis compares \emph{non-recursive} synthesis, in which every round samples from the original bootstrap seed pool, against \emph{recursive} synthesis, in which tasks that pass validation in the previous round become the seed pool for the next round.  All four groups share the generation model, rewrite operators, static checks, Daytona validation protocol, and repair budget, and each group is run for three synthesis rounds.  To control for the small and variable candidate counts in this pilot setting, we report a matched-subsample validation pass rate: for each configuration and round, we score the same size subsample of generated candidates against executable validation and report the fraction that pass.

\textbf{Solution-first generation and recursive seed continuation are each individually beneficial, and combining them yields the best configuration.}  Table~\ref{tab:2x2-ablation} reports the resulting pass rates.  Solution-first generation consistently outperforms instruction-first generation, with an average gain of $+24.1$ percentage points in the non-recursive setting and $+19.1$ in the recursive setting, indicating that constructing an executable reference solution and its environment before writing the public instruction produces tasks whose verifiers, instructions, and solutions remain internally consistent.  Recursive seed continuation likewise improves validation quality over the corresponding non-recursive baseline: it adds $+13.2$ points for instruction-first synthesis and $+8.2$ points for solution-first synthesis, with the strongest per-round gains appearing in $R_2$ and $R_3$, which is consistent with recursive seeds carrying more compatible task structure into later rounds than freshly sampled bootstrap seeds.  Combining both choices, solution-first recursive synthesis, the design used throughout the rest of the paper, achieves the highest average pass rate of $74.1\%$ and outperforms instruction-first non-recursive by $+32.3$ points, instruction-first recursive by $+19.1$ points, and solution-first non-recursive by $+8.2$ points.

\begin{table}[t]
\centering
\caption{$2\times 2$ ablation over generation order (instruction-first vs.\ solution-first) and seed-update strategy (non-recursive vs.\ recursive). Values are matched-subsample validation pass rates (\%) across three synthesis rounds ($R_1$--$R_3$). ``Avg'' is the mean pass rate across the three rounds, and the last three columns report absolute-percentage-point gains on the Avg column against the referenced baseline.}
\label{tab:2x2-ablation}
\small
\setlength{\tabcolsep}{4pt}
\renewcommand{\arraystretch}{1.18}
\resizebox{\textwidth}{!}{%
\begin{tabular}{lcccccccc}
\toprule
Group
& $R_1$
& $R_2$
& $R_3$
& Avg
& \shortstack{Gain vs.\\ IF non-rec.}
& \shortstack{Gain vs.\\ IF rec.}
& \shortstack{Gain vs.\\ SF non-rec.} \\
\midrule
Instruction-first, non-recursive          & 42.9 & 40.0 & 42.4 & 41.8 &  0.0 & $-13.2$ & $-24.1$ \\
Instruction-first, recursive              & 46.7 & 60.0 & 58.3 & 55.0 & $+13.2$ &  0.0 & $-10.9$ \\
Solution-first, non-recursive             & 66.7 & 64.3 & 66.7 & 65.9 & $+24.1$ & $+10.9$ &  0.0 \\
\midrule
\textbf{Solution-first, recursive (ours)} & \textbf{68.8} & \textbf{75.0} & \textbf{78.6} & \textbf{74.1} & \textbf{$+32.3$} & \textbf{$+19.1$} & \textbf{$+8.2$} \\
\bottomrule
\end{tabular}%
}
\end{table}

\subsection{Recursive Task Synthesis Results}
\label{sec:exp-synthesis-quality}

\textbf{Recursive synthesis remains stable across fifteen rounds rather than collapsing under repeated reuse.} Figure~\ref{fig:normalized-yield} reports two measures of synthesis stability: passed-task yield per 1,000 seed attempts and candidate pass rate. Passed-task yield remains between 498.2 and 572.2 across rounds, reaching 530.0 in $R_{15}$ compared with 551.6 in $R_1$. Candidate pass rate likewise stays within a narrow range, from 74.5\% to 81.5\%, with similar values in $R_1$ and $R_{15}$ (77.5\% and 78.0\%). These results show that the pipeline maintains comparable generation and validation performance even after fifteen rounds of recursive reuse.

\begin{figure*}[tbp]
  \centering
  \includegraphics[width=\textwidth]{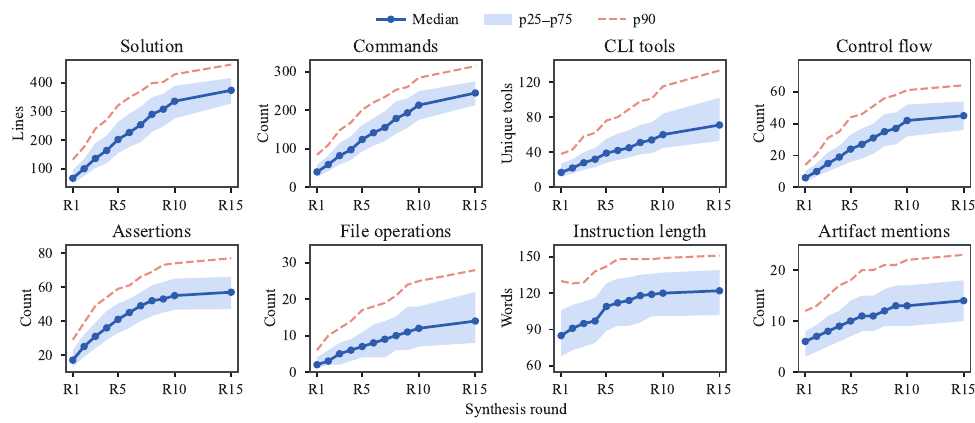}
  \caption{Quantile trends for eight task-structure metrics from $R_1$ to $R_{15}$.  Median and upper-quantile values increase most strongly for solution length, command use, CLI tools, control flow, assertions, and file operations, while instruction length grows comparatively slowly.}
  \label{fig:complexity-quantiles}
\end{figure*}

\begin{figure*}[!ht]
  \centering
  \begin{minipage}[t]{0.38\textwidth}
    \centering
    \includegraphics[width=\linewidth]{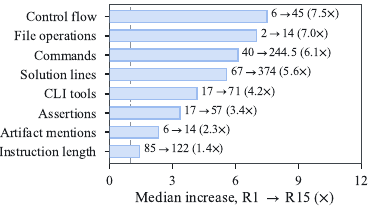}
    \caption{Median expansion factors from $R_1$ to $R_{15}$ across task-structure metrics.  Metrics associated with executable work increase substantially faster than instruction length.}
    \label{fig:r1-r10-gap}
  \end{minipage}\hfill
  \begin{minipage}[t]{0.60\textwidth}
    \centering
    \includegraphics[width=\linewidth]{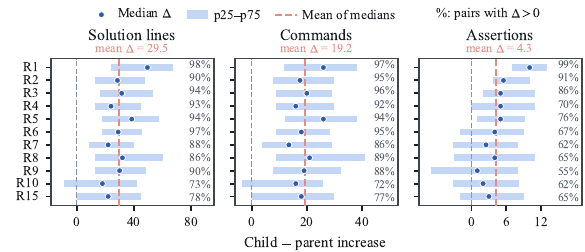}
    \caption{Parent-child changes in solution length, command count, and verifier assertions from $R_1$ to $R_{15}$.  Most accepted children exhibit positive changes relative to their immediate parents, indicating continued structural growth at the individual task level.}
    \label{fig:parent-child-delta}
  \end{minipage}
\end{figure*}

\textbf{At the same time, later-round tasks become substantially more complex in executable work rather than merely longer in prompt length.} Figures~\ref{fig:complexity-quantiles}--\ref{fig:parent-child-delta} show that from $R_1$ to $R_{15}$ the median solution length grows from 67 to 374 lines, command count from 40 to 244, unique CLI tools from 17 to 71, control-flow operations from 6 to 45, file operations from 2 to 14, and verifier assertions from 17 to 57, while instruction length grows much more slowly from 85 to 122 words. The upper quantiles rise in parallel, indicating that this growth is distributed across the task population rather than driven by a small number of outliers. Figure~\ref{fig:r1-r10-gap} summarizes the same pattern as expansion factors, and Figure~\ref{fig:parent-child-delta} confirms it at the task level: in $R_{15}$, median parent-child changes remain positive at 22 solution lines, 18 commands, and 3 verifier assertions, with positive deltas in 78\%, 77\%, and 65\% of pairs, respectively. Recursive synthesis therefore increases executable workload while continuing to add work at the individual task level.

\begin{figure*}[tbp]
  \centering
  \begin{minipage}[t]{0.55\textwidth}
    \centering
    \includegraphics[width=\linewidth]{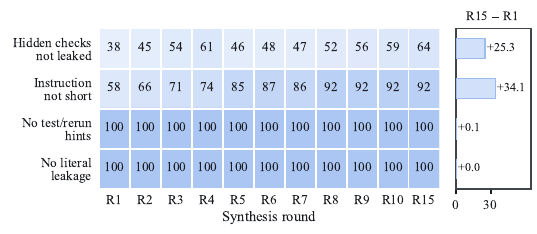}
    \caption{Public-instruction audit from $R_1$ to $R_{15}$. Hidden-check protection increases and short-instruction risk decreases, while references to private tests and literal leakage remain negligible.}
    \label{fig:alignment-quality}
  \end{minipage}\hfill
  \begin{minipage}[t]{0.43\textwidth}
    \centering
    \includegraphics[width=\linewidth]{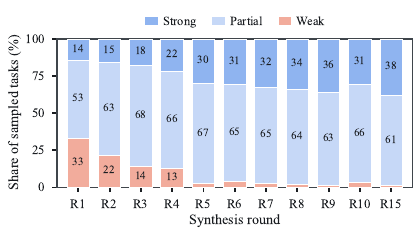}
    \caption{Requirement-to-verifier alignment from $R_1$ to $R_{15}$.  Sampled tasks are grouped by the proportion of public requirements reflected in executable checks, providing a direct measure of instruction-verifier consistency.}
    \label{fig:requirement-grounding}
  \end{minipage}
\end{figure*}

\textbf{Later-round tasks are not only more complex but also better aligned with their public instructions.} Figures~\ref{fig:alignment-quality} and~\ref{fig:requirement-grounding} show that hidden-check protection increases from 38.2\% to 63.5\%, while short-instruction risk decreases from 41.6\% to 7.5\%; references to private tests and literal leakage remain negligible, reaching only 0.1\% and 0\% in $R_{15}$, respectively. Median requirement coverage rises from 0.42 to 0.57, the share of weakly grounded tasks falls from 32.8\% to 1.2\%, and the strongly grounded share increases from 14.3\% to 38.0\%. Together, these results indicate that later tasks are better specified and that both binary success and partial-credit scores more faithfully reflect the public task contract.

\subsection{Task Difficulty Analysis}
\label{sec:exp-fixed-model-difficulty}

To test whether the task-side growth observed above translates into agent-side difficulty, we evaluate DeepSeek-V4-Pro on matched task subsets from each synthesis round under fixed inference settings. Pass@4 measures full task completion within four attempts, while partial credit measures the fraction of verifier checks satisfied.

\begin{figure*}[tbp]
  \centering
  \begin{minipage}[t]{0.242\textwidth}
    \centering
    \includegraphics[width=\linewidth]{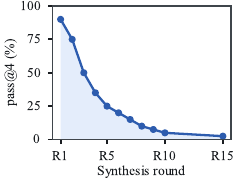}
    \caption{DeepSeek-V4-Pro pass@4 on task subsets across recursive synthesis rounds. The success decreases monotonically from 90\% in $R_1$ to 2.5\% in $R_{15}$.}
    \label{fig:deepseek-pass4}
  \end{minipage}\hfill
  \begin{minipage}[t]{0.242\textwidth}
    \centering
    \includegraphics[width=\linewidth]{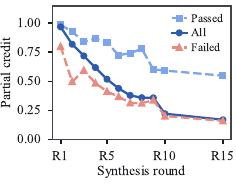}
    \caption{Mean verifier-based partial credit achieved by DeepSeek-V4-Pro, declining from 0.970 in $R_1$ to 0.170 in $R_{15}$.}
    \label{fig:deepseek-process}
  \end{minipage}\hfill
  \begin{minipage}[t]{0.242\textwidth}
    \centering
    \includegraphics[width=\linewidth]{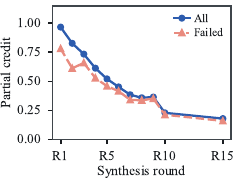}
    \caption{Attempt-level partial credit from $R_1$ to $R_{15}$, for all attempts and for failed attempts.}
    \label{fig:deepseek-attempt-process}
  \end{minipage}\hfill
  \begin{minipage}[t]{0.242\textwidth}
    \centering
    \includegraphics[width=\linewidth]{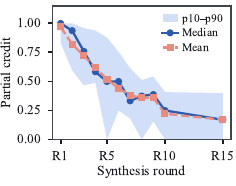}
    \caption{Task-level partial credit distribution, with median and p10--p90 shifting downward across rounds.}
    \label{fig:deepseek-process-distribution}
  \end{minipage}
\end{figure*}

\textbf{Later-round tasks become substantially harder for a fixed solver.} Figure~\ref{fig:deepseek-pass4} shows that DeepSeek-V4-Pro pass@4 declines monotonically from 90\% in $R_1$ to 2.5\% in $R_{15}$, a 36-fold reduction in full-task success. Figure~\ref{fig:deepseek-process} shows that mean partial credit falls in parallel, from 0.970 to 0.170. Because the solver and inference configuration remain unchanged across rounds, this decline reflects changes in the task distribution rather than changes in model capability. The results therefore show that the structural growth produced by recursive synthesis corresponds to genuinely greater task difficulty.

\textbf{This increase in difficulty is broad rather than being driven by aggregation artifacts or a small number of outliers.} Figure~\ref{fig:deepseek-attempt-process} shows that mean attempt-level partial credit decreases from 0.968 in $R_1$ to 0.180 in $R_{15}$, while the mean for failed attempts drops from 0.782 to 0.160, indicating that reduced progress appears consistently across individual rollouts. Figure~\ref{fig:deepseek-process-distribution} further shows that the entire task-level partial-credit distribution shifts downward: the median falls from 1.0 to 0.170, the p10 reaches 0, and the p90 declines to 0.400 by $R_{15}$. Increased difficulty is therefore distributed broadly across later-round tasks rather than concentrated in a narrow upper tail.

\begin{figure*}[tbp]
  \centering
  \begin{minipage}[t]{0.325\textwidth}
    \centering
    \includegraphics[width=\linewidth]{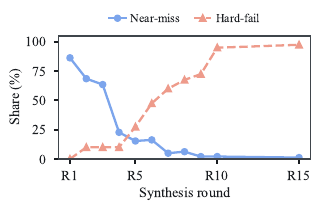}
    \caption{Distributional shift in partial-credit outcomes.  Failed attempts above 0.75 partial credit fall from 86.4\% to 1.2\%, while tasks below 0.50 rise from 0\% to 97.5\%.}
    \label{fig:deepseek-nearmiss-hardfail}
  \end{minipage}\hfill
  \begin{minipage}[t]{0.325\textwidth}
    \centering
    \includegraphics[width=\linewidth]{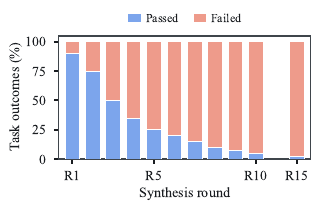}
    \caption{Task-level pass@4 outcomes for DeepSeek-V4-Pro.  The pass rate decreases from 90\% in $R_1$ to 20\% in $R_6$ and 2.5\% in $R_{15}$.}
    \label{fig:deepseek-pass-fail-share}
  \end{minipage}\hfill
  \begin{minipage}[t]{0.325\textwidth}
    \centering
    \includegraphics[width=\linewidth]{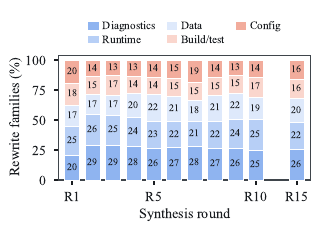}
    \caption{Distribution of accepted tasks across the five rewrite families.  Entropy stays close to its maximum, and no family exceeds 29\% in any round.}
    \label{fig:family-diversity}
  \end{minipage}
\end{figure*}

\textbf{Later-round failures are also qualitatively harder, with far fewer near-misses and little recovery from repeated sampling.} Figure~\ref{fig:deepseek-nearmiss-hardfail} shows that failed attempts satisfying at least 75\% of verifier checks fall from 86.4\% in $R_1$ to 1.2\% in $R_{15}$, while the share of tasks below half of the checks rises from 0\% to 97.5\%. Figure~\ref{fig:deepseek-pass-fail-share} shows that DeepSeek-V4-Pro solves at most 15\% of tasks from $R_7$ onward, with pass@4 falling to 2.5\% in $R_{15}$. Together, these results show that recursive synthesis does not merely reduce binary success rates; it produces later-round tasks on which the same solver makes substantially less partial progress and recovers far fewer complete solutions.

\subsection{Data Diversity and Duplication Analysis}
\label{sec:exp-diversity-novelty}

\textbf{Recursive synthesis preserves broad diversity across domains and rewrite families rather than collapsing onto a single mode of task generation.} Figure~\ref{fig:seed-domain-dist} shows that the task pool retains broad domain coverage across synthesis rounds, while Figure~\ref{fig:family-diversity} shows that rewrite-family entropy remains between 2.26 and 2.31 bits, close to the maximum of $\log_2 5 = 2.32$ bits for five families. The largest family accounts for only 24.9\% to 28.9\% of accepted tasks across rounds, and the $R_{15}$ distribution remains well spread across diagnostics and forensics, runtime substrate, data and artifact processing, build and test workflows, and configuration and state migration. The structural growth reported in Figures~\ref{fig:complexity-quantiles}--\ref{fig:parent-child-delta} therefore reflects multiple forms of task extension rather than one dominant rewrite pattern.

\begin{figure*}[tbp]
  \centering
  \begin{minipage}[t]{0.52\textwidth}
    \centering
    \includegraphics[width=\linewidth]{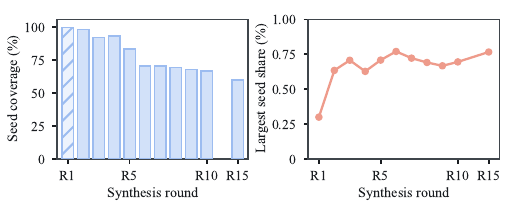}
    \caption{Coverage of bootstrap seeds across recursive rounds. $R_{15}$ retains 218 of the 363 seeds represented in $R_1$, and no single seed contributes more than 0.77\% of the $R_{15}$ tasks.}
    \label{fig:parent-coverage}
  \end{minipage}\hfill
  \begin{minipage}[t]{0.46\textwidth}
    \centering
    \includegraphics[width=\linewidth]{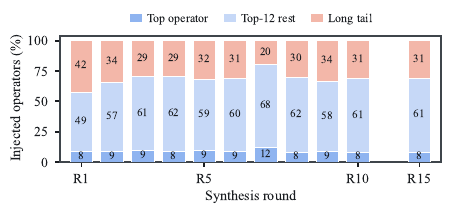}
    \caption{Rewrite-operator coverage from $R_1$ to $R_{15}$. Each round uses 31 to 38 of the 40 available operators, and the most frequent operator accounts for at most 12.2\% of the tasks.}
    \label{fig:operator-long-tail}
  \end{minipage}
\end{figure*}

\textbf{Repeated reseeding does not cause the dataset to become dominated by a small set of parents or operators.} Figure~\ref{fig:parent-coverage} shows that bootstrap-seed coverage decreases gradually from 98.3\% in $R_2$ to 60.1\% in $R_{15}$, indicating that some seed sources are filtered out over time. However, no individual seed contributes more than 0.77\% of the tasks in any round, so the remaining pool does not become concentrated around a small number of lineages. Figure~\ref{fig:operator-long-tail} likewise shows that the pipeline continues to use a broad set of rewrite operations: 36 of the 40 operators remain represented in $R_{15}$, the most frequent operator accounts for only 8.0\% of tasks, and operators outside the twelve most frequent still account for 31.0\%. Recursive reuse therefore remains diverse both in ancestry and in transformation mechanism.

\begin{figure*}[tbp]
  \centering
  \begin{minipage}[t]{0.55\textwidth}
    \centering
    \includegraphics[width=\linewidth]{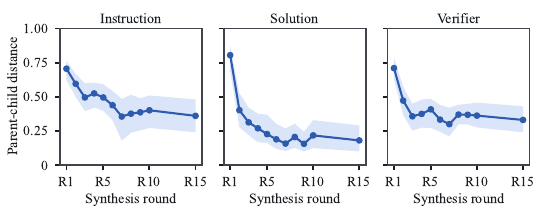}
    \caption{Token-level parent-child novelty in the instruction, solution, and verifier from $R_1$ to $R_{15}$.  Later-round rewrites retain more content from their parents while continuing to modify all three task components.}
    \label{fig:parent-child-novelty}
  \end{minipage}\hfill
  \begin{minipage}[t]{0.43\textwidth}
    \centering
    \includegraphics[width=\linewidth]{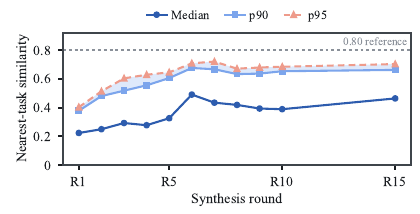}
    \caption{Within-round nearest-neighbor similarity from $R_1$ to $R_{15}$. The $R_{15}$ median remains below 0.5, while p95 reaches 0.703, indicating broad task variation with a limited high-similarity tail.}
    \label{fig:near-duplicate-risk}
  \end{minipage}
\end{figure*}

\textbf{Although later-round tasks become somewhat more similar, recursive synthesis does not collapse into repeated copies.} Figures~\ref{fig:parent-child-novelty} and~\ref{fig:near-duplicate-risk} show that in $R_{15}$ the median parent-child novelty remains 0.36 for the instruction, 0.18 for the solution, and 0.33 for the verifier, indicating that later children still modify all three task components even while retaining more implementation content from their parents. Within-round nearest-neighbor similarity increases from a median of 0.223 in $R_1$ to 0.464 in $R_{15}$, but the median remains below 0.5, meaning that a typical $R_{15}$ task still shares fewer than half of its combined instruction, solution, and verifier tokens with its closest neighbor. The p95 value of 0.703 shows that high similarity is concentrated in a limited upper tail rather than across the entire task pool. Taken together, these results show that recursive synthesis increases difficulty without collapsing into duplicates, while also highlighting a manageable high-similarity tail for future deduplication.

\subsection{Supervised Fine-Tuning Results}
\label{sec:exp-downstream}

The preceding analyses show that recursive synthesis produces increasingly difficult tasks while maintaining validation stability and broad data coverage. We next evaluate whether trajectories collected from these tasks improve terminal-agent performance. Successful Qwen3.5 rollouts are used to fine-tune Qwen3.5-27B and Qwen3.5-122B-A10B. The trained checkpoints are evaluated on Terminal-Bench~2~\citep{merrill2026terminalbenchbenchmarkingagentshard}, Terminal-Bench Hard~\citep{ivison2026tmaxsimplerecipeterminal}, and Long-Horizon Terminal Bench~\citep{li2026longhorizonterminalbenchtestinglimitsagents}.  Each checkpoint is compared with its corresponding base model under the same evaluation configuration, allowing to measure both the overall training gain and the effect of incorporating additional recursive rounds.

\begin{figure*}[tbp]
  \centering
  \begin{minipage}[t]{0.55\textwidth}
    \centering
    \includegraphics[width=\linewidth]{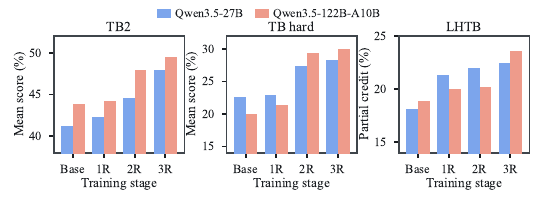}
    \caption{Benchmark performance of Qwen3.5-27B and Qwen3.5-122B-A10B after supervised fine-tuning on trajectories from progressively more synthesis rounds, reported on Terminal-Bench~2, Terminal-Bench Hard, and Long-Horizon Terminal Bench (mean partial credit).}
    \label{fig:qwen-round-benchmarks}
  \end{minipage}\hfill
  \begin{minipage}[t]{0.43\textwidth}
    \centering
    \includegraphics[width=\linewidth]{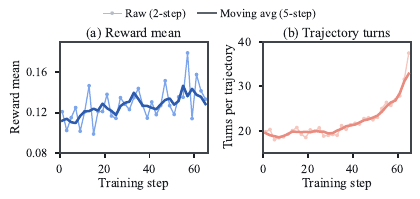}
    \caption{Reinforcement-learning dynamics on the synthesized terminal tasks: (a) mean verifier reward, rising from approximately 0.11 to above 0.14, and (b) mean interaction turns per trajectory, rising from 19--20 to more than 30 turns.}
    \label{fig:rl-dynamics}
  \end{minipage}
\end{figure*}

\begin{table*}[t]
\centering
\small
\setlength{\tabcolsep}{10pt}
\renewcommand{\arraystretch}{1.15}
\caption{Evaluation of Qwen3.5-27B and Qwen3.5-122B-A10B across successive training rounds on Terminal-Bench~2 (TB2), Terminal-Bench Hard, and the Long-Horizon Terminal Benchmark (LHTB). Base denotes the checkpoint before training, while Round~1--3 denote checkpoints obtained after successive training rounds. Each entry reports the evaluation mean $\pm$ standard deviation.}
\label{tab:qwen35_round_results}
\begin{tabular}{llccc}
\toprule
\multirow{2}{*}{Model (Non-thinking Mode) }
& \multirow{2}{*}{Training stage}
& \multicolumn{3}{c}{Pass rate (\%, mean $\pm$ std.)} \\
\cmidrule(lr){3-5}
& & Terminal-Bench 2  & Terminal-Bench~Hard & LHTB \\
\midrule
\multirow{4}{*}{Qwen3.5-27B}
& Base    & $41.20 \pm 1.72$ & $22.67 \pm 2.52$ & $18.10 \pm 0.89$ \\
& Round 1 & $42.32 \pm 4.68$ & $23.00 \pm 1.73$ & $21.32 \pm 0.86$ \\
& Round 2 & $44.57 \pm 4.25$ & $27.33 \pm 1.53$ & $21.99 \pm 1.39$ \\
& Round 3 & $47.94 \pm 2.34$ & $28.33 \pm 0.58$ & $22.44 \pm 0.40$ \\
\midrule
\multirow{4}{*}{Qwen3.5-122B-A10B}
& Base    & $43.82 \pm 2.97$ & $20.00 \pm 1.00$ & $18.85 \pm 1.16$ \\
& Round 1 & $44.19 \pm 1.30$ & $21.33 \pm 3.21$ & $20.06 \pm 0.32$ \\
& Round 2 & $47.94 \pm 1.72$ & $29.33 \pm 2.08$ & $20.24 \pm 0.60$ \\
& Round 3 & $49.44 \pm 1.12$ & $30.00 \pm 1.73$ & $23.63 \pm 2.76$ \\
\bottomrule
\end{tabular}
\end{table*}

Figure~\ref{fig:qwen-round-benchmarks} and Table~\ref{tab:qwen35_round_results} compare the base models with checkpoints fine-tuned on trajectories from one, two, and three recursive synthesis stages.  Performance increases monotonically for both model sizes on all three benchmarks.  At the three-round stage, Terminal-Bench~2 improves from 41.2\% to 47.9\% for Qwen3.5-27B and from 43.8\% to 49.4\% for Qwen3.5-122B-A10B.  Terminal-Bench Hard improves from 22.7\% to 28.3\% and from 20.0\% to 30.0\%, respectively. On Long-Horizon Terminal Bench, mean partial credit increases from 18.1\% to 22.4\% for Qwen3.5-27B and from 18.9\% to 23.6\% for Qwen3.5-122B-A10B.

All checkpoints are evaluated under the same benchmark configurations and compared with the corresponding base model. The consistent gains across training stages show that adding trajectories from further synthesis rounds continues to improve performance beyond the initial round. Improvements on both model sizes indicate that the benefit is not specific to a single model scale. Performance gains on Terminal-Bench~2, Terminal-Bench Hard, and Long-Horizon Terminal Bench demonstrate transfer across broad terminal execution, independently constructed tasks, and long-horizon interaction.

\subsection{Comparison Against Public Terminal-Agent SFT Datasets Under Matched Supervision Budget}
\label{sec:exp-sft-dataset-comparison}

The preceding SFT results show that trajectories collected on the recursively synthesized task pool transfer to downstream terminal-agent benchmarks, but they do not isolate the contribution of the underlying task pool from the general effect of terminal-agent supervision.  To directly compare our synthesized trajectories against other publicly released terminal-agent SFT datasets, we conduct a controlled experiment that fixes the total supervision budget rather than the number of trajectories.  Trajectory lengths differ substantially across public datasets, so matching by trajectory count would give each run a very different amount of gradient-relevant supervision; matching by \emph{assistant-loss tokens}, i.e., the tokens that actually participate in the supervised loss, fixes the true amount of information supplied to the model.

\textbf{Data setting.}  All six experiments use the same 36{,}363{,}275 assistant-loss tokens, corresponding to the full supervision budget of RST-only's 11{,}531 unique trajectories.  The compared datasets are OpenThoughts-Agent-v1-SFT~\citep{raoof2026openthoughtsagentdatarecipesagentic}, TMax-SFT~\citep{ivison2026tmaxsimplerecipeterminal}, LiteCoder-Terminal-SFT~\citep{peng2026litecoderterminalscalinglonghorizonterminal}, Nemotron-Terminal-Corpus (skill-based subsets)~\citep{pi2026dataengineeringscalingllm}, SETA-SFT~\citep{shen2026setascalingenvironmentsterminal}, and our RST-only pool.  For Nemotron-Terminal-Corpus we use only the \texttt{skill\_based\_easy}, \texttt{skill\_based\_medium}, and \texttt{skill\_based\_mixed} subsets, and for SETA we use the 1{,}112 reward-1 successful trajectories reported in the original paper.  Datasets whose unique-trajectory tokens do not reach the target budget in a single pass are looped, reshuffled, and resampled until the budget is exactly matched, so the ``trajectory exposures'' column in Table~\ref{tab:sft-dataset-comparison} differs from the ``unique trajectories'' column for those datasets.  Random sampling uses seed 42, and tokens beyond the budget within the last selected trajectory are masked out (\texttt{loss\_mask}$=0$) so all six runs terminate at exactly the same supervision budget.

\textbf{Preprocessing.}  All data are converted to Qwen3.5 messages format using the Qwen3.5-27B tokenizer, the Qwen3.5 chat template, and the SLIME \texttt{qwen3\_5} loss mask.  Only assistant reasoning, textual replies, and tool calls receive gradient; system, user, and tool-observation tokens are masked out.  Existing reasoning content and tool calls in the source data are preserved.  The maximum context length is 262{,}145 tokens; trajectories exceeding this limit are truncated under a shared rule before assistant-loss tokens are counted.  Before training we apply joint benchmark decontamination against the three evaluation sets across all six datasets and resample from the remaining data to restore the target budget.

\textbf{Training configuration.}  All six checkpoints are independently initialized from the same Qwen3.5-27B Base checkpoint and trained with full-parameter SFT, without cross-dataset continued training and without mixing auxiliary data.  Optimization uses Adam with peak learning rate $3\times 10^{-6}$, minimum learning rate $3\times 10^{-7}$, cosine decay, 5\% warmup, betas $(0.9, 0.98)$, weight decay $0$, dropout $0$, and BF16 precision.  Global batch size is $128$, and dynamic batching and sequence packing are kept identical across runs.  The learning-rate schedule and stopping condition are keyed to cumulative assistant-loss tokens rather than epochs, so that the supervision budget stays fixed even when datasets differ in trajectory count or repetition factor.  This design controls the number of supervision-target tokens rather than total context tokens or realized training FLOPs.

\textbf{Evaluation configuration.}  All checkpoints are evaluated with a shared Terminus-2, Harbor, Daytona, and SGLang configuration on Terminal-Bench~2, Terminal-Bench Hard, and Long-Horizon Terminal Bench.  Inference uses context length $131{,}072$, temperature $0.6$, top-$p$ $0.95$, top-$k$ $20$, at most $8{,}192$ output tokens per call, and non-thinking mode.  Maximum turns are $100$ on Terminal-Bench~2 and Terminal-Bench Hard and $1{,}000$ on Long-Horizon Terminal Bench.  Each checkpoint is evaluated three times per benchmark, and we report mean $\pm$ standard deviation, where the standard deviation measures variation across the three independent runs and improvements are computed against the same-configuration Base model.

\textbf{Results.}  Table~\ref{tab:sft-dataset-comparison} reports the resulting scores.  Fine-tuning on RST-only trajectories yields the strongest result on every benchmark: $46.82\%$ on Terminal-Bench~2, $27.67\%$ on Terminal-Bench Hard, and $21.90\%$ on Long-Horizon Terminal Bench, corresponding to absolute improvements of $+5.62$, $+5.00$, and $+3.80$ percentage points over the Qwen3.5-27B base model.  Under the same supervision budget, the best public terminal SFT dataset in this comparison is LiteCoder-Terminal-SFT, which reaches $42.70\%$, $24.33\%$, and $20.00\%$, so RST-only exceeds the strongest public baseline by $+4.12$, $+3.34$, and $+1.90$ points across the three benchmarks.  OpenThoughts-Agent, TMax-SFT, and SETA-SFT produce only marginal gains over the base model (at most $+1.5$ points on TB2 and $+1.2$ on LHTB), and Nemotron-Terminal-Corpus (skill-based) actually decreases TB2 and TB-Hard performance by $1.50$ and $1.00$ points at the same supervision budget, suggesting that its skill-focused single-turn format transfers poorly to full terminal-agent trajectories.  These results indicate that the improvement observed with RST-only reflects the underlying task pool rather than a generic effect of terminal-agent supervision.

\begin{table*}[t]
\centering
\caption{Comparison of six terminal-agent SFT datasets under a matched supervision budget of 36{,}363{,}275 assistant-loss tokens, corresponding to the full budget of RST-only's 11{,}531 unique trajectories.  For each dataset we report the number of unique trajectories in the pool used, the effective number of trajectories fed to training (including repetitions for pools smaller than the target budget), and the resulting Terminal-Bench~2, Terminal-Bench Hard, and Long-Horizon Terminal Bench scores (\%) as mean $\pm$ standard deviation over three independent evaluation runs.  Trajectories are drawn with random seed~42, and tokens beyond the budget within the last selected trajectory are set to \texttt{loss\_mask}$=0$ so all six SFT runs terminate at exactly the same supervision budget.  The Qwen3.5-27B Base row reports the same-configuration base model as a no-SFT reference.}
\label{tab:sft-dataset-comparison}
\small
\setlength{\tabcolsep}{4pt}
\renewcommand{\arraystretch}{1.22}
\resizebox{\textwidth}{!}{%
\begin{tabular}{lrrrccc}
\toprule
Training data
& \shortstack{Unique\\ trajectories}
& \shortstack{Trajectory\\ exposures}
& \shortstack{Assistant-loss\\ tokens}
& TB2 (Mean $\pm$ Std.)
& TB-Hard (Mean $\pm$ Std.)
& LHTB (Mean $\pm$ Std.) \\
\midrule
Qwen3.5-27B Base (no SFT)
& --- & --- & --- & $41.20 \pm 1.72$ & $22.67 \pm 1.53$ & $18.10 \pm 0.80$ \\
\midrule
OpenThoughts-Agent-v1-SFT~\citep{raoof2026openthoughtsagentdatarecipesagentic}
& 15{,}209 & 15{,}613 & 36{,}363{,}275 & $41.95 \pm 1.30$ & $23.67 \pm 0.58$ & $19.30 \pm 1.05$ \\

TMax-SFT~\citep{ivison2026tmaxsimplerecipeterminal}
& 10{,}726 & 8{,}314 & 36{,}363{,}275 & $41.57 \pm 1.12$ & $23.67 \pm 1.15$ & $19.10 \pm 0.75$ \\

LiteCoder-Terminal-SFT~\citep{peng2026litecoderterminalscalinglonghorizonterminal}
& 11{,}255 & 3{,}071 & 36{,}363{,}275 & $42.70 \pm 1.12$ & $24.33 \pm 2.08$ & $20.00 \pm 1.05$ \\

Nemotron-Terminal-Corpus (skill-based)~\citep{pi2026dataengineeringscalingllm}
& 139{,}841 & 6{,}272 & 36{,}363{,}275 & $39.70 \pm 0.65$ & $21.67 \pm 0.58$ & $18.30 \pm 0.72$ \\

SETA-SFT~\citep{shen2026setascalingenvironmentsterminal}
& 1{,}112 & 6{,}991 & 36{,}363{,}275 & $40.45 \pm 1.12$ & $22.67 \pm 1.15$ & $18.80 \pm 0.85$ \\

\midrule
\textbf{RST-only (ours)}
& \textbf{11{,}531} & \textbf{11{,}531} & \textbf{36{,}363{,}275}
& $\mathbf{46.82 \pm 1.30}$ & $\mathbf{27.67 \pm 1.53}$ & $\mathbf{21.90 \pm 1.00}$ \\

\bottomrule
\end{tabular}%
}

\vspace{0.3em}
\begin{minipage}{\textwidth}
\footnotesize
Nemotron-Terminal-Corpus is restricted to the \texttt{skill\_based\_easy}, \texttt{skill\_based\_medium}, and \texttt{skill\_based\_mixed} subsets released with the corpus.  SETA-SFT uses the 1{,}112 reward-1 successful trajectories reported in the original paper.  OpenThoughts-Agent and SETA are looped and reshuffled to reach the supervision-budget target because their unique-trajectory tokens fall short of the target in one pass; the ``trajectory exposures'' column therefore differs from the ``unique trajectories'' count for these datasets.
\end{minipage}
\end{table*}

\subsection{Terminal Agentic Reinforcement Learning Results}
\label{sec:exp-rl-dynamics}

We then report our agentic reinforcement-learning dynamics on the synthesized tasks.  The policy is initialized from the Qwen3.5-27B base checkpoint, corresponding to a cold actor start from the released base weights at rollout step zero.  The PPO value head is warm-loaded from a prior terminal-agent critic checkpoint and receives two critic-only warm-up steps.
The optimizer uses PPO with an actor--critic value function and generalized advantage estimation.  The KL penalty and entropy bonus are disabled, PPO clipping uses $\epsilon=0.2$, and advantages are whitened to zero mean and unit variance per batch.

The RL training pool is the synth-all set of 37,484 synthesized terminal-agent tasks, consisting of the $R_1$ synthesized pool and oracle-passed tasks from $R_2$--$R_{15}$ of the self-improvement loop.  Rewrite variants are retained without deduplication.  Each task is a self-contained terminal environment with task metadata, public instruction, verifier tests, reference solution, and Dockerfile, executed in a Daytona sandbox.  The pool is reshuffled each epoch so that each batch mixes rounds and difficulty levels. Reward is computed from each task's built-in verifier with customized reward shaping.
Figure~\ref{fig:rl-dynamics} shows raw measurements logged at two-step intervals together with a five-step moving average.

Figure~\ref{fig:rl-dynamics}(a) reports mean verifier reward during PPO training.  Reward varies across updates because each batch contains tasks with different difficulty levels and numbers of verifier checks.  Nevertheless, the five-step moving average increases from approximately 0.11 at the beginning of training to above 0.14 in later updates, with its highest values occurring around steps 55 to 60.  The increase indicates that the policy satisfies more verifier checks as training progresses and that the synthesized tasks provide an effective graded learning signal. Figure~\ref{fig:rl-dynamics}(b) shows that mean trajectory length also increases during training.  The five-step moving average remains near 19 to 20 turns during the early updates, begins rising after approximately step 30, and exceeds 30 turns in the later stage.  This increase occurs alongside the rise in verifier reward, indicating that longer interactions are associated with greater task progress rather than an immediate decline in execution quality.  The policy therefore learns to sustain longer terminal interactions while satisfying more verifier checks.

\begin{table}[H]
  \centering
  \caption{Benchmark performance of DeepSeek-V4-Pro, Qwen3.5-27B Base, Qwen3.5-122B-A10B Base, and Qwen3.5-27B-RL on Terminal-Bench~2, Terminal-Bench Hard, and Long-Horizon Terminal Bench. All scores are averaged over three independent evaluation runs. The final row reports the relative performance gain of Qwen3.5-27B-RL over Qwen3.5-27B Base.}
  \label{tab:rl-benchmark-comparison}
  \small
  \resizebox{0.64\linewidth}{!}{%
    \begin{tabular}{lccc}
      \toprule
      Model (Non-thinking Mode) & Terminal-Bench~2 & Terminal-Bench~Hard & LHTB \\
      \midrule
      Deepseek-V4-Pro & 51.68 & 36.00 & 30.00 \\
      Qwen3.5-27B Base & 41.20 & 22.67 & 18.10 \\
      Qwen3.5-122B-A10B Base & 43.82 & 20.00 & 18.85 \\
      Qwen3.5-27B-RL & 49.44 & 32.00 & 22.07 \\
      \midrule
      Relative gain & +20.00\% & +41.16\% & +21.93\% \\
      \bottomrule
    \end{tabular}%
  }
\end{table}

Table~\ref{tab:rl-benchmark-comparison} reports mean performance over three evaluation runs. Qwen3.5-27B-RL reaches 49.44\% on Terminal-Bench~2, 32.00\% on Terminal-Bench Hard, and 22.07\% on Long-Horizon Terminal Bench.  Relative to Qwen3.5-27B Base, these results correspond to improvements of 20.00\%, 41.16\%, and 21.93\%, respectively. The largest relative gain occurs on Terminal-Bench Hard, indicating that verifier-based RL transfers to tasks constructed independently from the training pool. DeepSeek-V4-Pro remains stronger across all three benchmarks, showing that substantial performance headroom remains.

\subsection{Case Study: Recursive Growth of a JSON-Diff Regression Task}
\label{sec:case-study-gendiff}

To illustrate how a task evolves across rounds, we follow one lineage from the bootstrap seed to $R_{15}$. The original task asks the agent to run the \texttt{gendiff} command-line tool on fixed pairs of JSON fixtures and save the generated reports. Later rounds retain this objective but add configuration files, regression cases, failure diagnosis, release-note updates, and executable tests.

Table~\ref{tab:gendiff-case-study} shows five checkpoints in this lineage. The added difficulty comes from new executable requirements rather than longer instructions. Later versions introduce a task matrix in \texttt{tasks.json}, checks for invalid configuration, a new regression case, and consistency constraints across fixtures, expected counts, reports, and tests. By $R_{15}$, the agent must repair the configuration and fixture data, regenerate all reports, reconcile observed and expected counts, and pass the project test suite.

\begin{table*}[t]
\centering
\footnotesize
\setlength{\tabcolsep}{3.5pt}
\renewcommand{\arraystretch}{1.22}
\caption{Case study of one exact recursive lineage. The task keeps the same core objective, generating reliable JSON-diff regression outputs, while later rounds add configuration, failure diagnosis, release-note updates, and count reconciliation. The last column reports the number of non-empty lines in the reference \texttt{solution/solve.sh}.}
\label{tab:gendiff-case-study}
\begin{tabular}{@{}>{\centering\arraybackslash}m{0.08\textwidth}
                  m{0.24\textwidth}
                  m{0.23\textwidth}
                  m{0.32\textwidth}
                  >{\centering\arraybackslash}m{0.05\textwidth}@{}}
\toprule
\textbf{Checkpoint} &
\textbf{Public task} &
\textbf{Added context} &
\textbf{Agent work required} &
\textbf{Lines} \\
\midrule
Seed &
Generate six JSON-diff reports from fixed fixture pairs. &
Manual CLI example with fixed input files. &
Run \texttt{gendiff} on two JSON fixture pairs in three output formats. &
10 \\
R1 &
Generate all diff outputs specified by \texttt{tasks.json}. &
The comparison matrix is moved into configuration. &
Read \texttt{tasks.json}, batch-run all comparisons, and write a validation report. &
66 \\
R5 &
Repair \texttt{tasks.json} using \texttt{check\_config.sh}, then regenerate the diffs. &
The configuration may be invalid and must be diagnosed. &
Run the checker, interpret failures, fix the task list, rerun the suite, and aggregate change counts. &
219 \\
R10 &
Use \texttt{CHANGELOG.md} to add the new nested-fixture regression case. &
A release note introduces a new fixture and required comparison. &
Update the regression matrix, generate the new output, and satisfy the local report validator. &
264 \\
R15 &
Repair inconsistent configuration and fixture data; reconcile expected diff counts. &
Configuration, fixtures, expected counts, and tests must agree. &
Fix \texttt{tasks.json} and JSON fixtures, regenerate every diff, write diagnosis and validation reports, and pass tests. &
347 \\
\bottomrule
\end{tabular}
\end{table*}

This lineage provides a concrete example of the complexity growth shown in Figures~\ref{fig:complexity-quantiles}--\ref{fig:parent-child-delta}. The core objective remains unchanged, but each round adds more executable work. The agent must inspect workspace evidence, diagnose inconsistent state, repair configuration and data, regenerate artifacts, and pass executable tests. As a result, the reference solution grows from 10 to 347 non-empty shell lines without changing the underlying task domain.

\section{Conclusion}
\label{sec:conclusion}

We presented \ours{}, a framework for recursively constructing verified terminal-agent tasks.  Each synthesis round extends the reference solution to introduce additional executable work, updates the verifier and public instruction to describe the same task, and validates the complete candidate in a fresh sandbox.  A task is accepted only when its reference solution passes the private verifier and every tested requirement is stated in the instruction or discoverable from the workspace.  Accepted tasks seed subsequent synthesis rounds and directly form the task pool for verifier-based reinforcement learning.  Successful rollouts collected from the same tasks provide trajectories for supervised fine-tuning.

Across fifteen synthesis rounds, \ours{} produces 37,484 verified tasks at a cost of approximately \$50 per 1,000 accepted tasks while maintaining stable validation yield.  From $R_1$ to $R_{15}$, median solution length increases by 5.6$\times$ and command use by 6.1$\times$, while instruction length increases by only 1.4$\times$.  DeepSeek-V4-Pro pass@4 decreases from 90\% to 2.5\%, and mean partial credit decreases from 0.970 to 0.170, confirming that later tasks require substantially more agent capability.  Supervised fine-tuning improves both Qwen3.5-27B and Qwen3.5-122B-A10B across Terminal-Bench~2, Terminal-Bench Hard, and Long-Horizon Terminal Bench.  Qwen3.5-27B-RL reaches 49.44\%, 32.00\%, and 22.07\% on the three benchmarks, corresponding to relative gains of 11.82\%, 41.16\%, and 21.93\% over the base model.  Domain, rewrite-family, and operator coverage remain broad across rounds, although the high-similarity tail motivates targeted deduplication during further scaling.

\bibliographystyle{plainnat}
\bibliography{refs}

\newpage
\appendix
\section{Bootstrap Seed Pool Diversity}
\label{app:seed-diversity}

Section~\ref{sec:method-seeds} describes the 639 oracle-valid bootstrap tasks, whose high-level domain distribution is shown in Figure~\ref{fig:seed-domain-dist}.  We consolidate the original fine-grained category labels into broader domains for analysis.  Table~\ref{tab:seed-domain-mapping} provides the complete mapping between the original categories and the resulting analytical groupings.

To further expand the diversity of our bootstrap seed pool, future work could incorporate tasks from additional scientific and engineering domains, such as medical image segmentation~\cite{ronneberger2015u,zhou2018unet++,li2021pneumothorax}, environmental modeling and wildfire prediction~\cite{reichstein2019deep,li2021wildland}, and intelligent fault diagnosis~\cite{lei2020applications,li2023lightweight}. These domains offer opportunities to develop new seed tasks with diverse data formats, computational workflows, and domain-specific verification requirements.

\begin{sidewaystable*}
\centering
\small
\setlength{\tabcolsep}{5pt}
\caption{Mapping from the high-level domain labels in Figure~\ref{fig:seed-domain-dist} to the original categories for the 639 oracle-valid bootstrap tasks. Domains are post-hoc merges of overlapping raw categories; they are not native dataset fields.}
\label{tab:seed-domain-mapping}
\begin{tabular}{lrp{0.72\textwidth}}
\toprule
Domain & \#Tasks & Original TerminalWorld categories \\
\midrule
Build \& Compile & 149 & \texttt{build-system}, \texttt{software-development}, \texttt{software-engineering} \\
Scripting \& Automation & 103 & \texttt{scripting}, \texttt{scripting-automation} \\
System Admin \& Services & 87 & \texttt{process-management}, \texttt{system-administration} \\
Security / Reverse / Crypto & 52 & \texttt{security} \\
Package \& Environment & 47 & \texttt{environment-setup}, \texttt{package-management} \\
Version Control \& Release & 34 & \texttt{version-control} \\
Containers \& Kubernetes & 34 & \texttt{container-cloud-native}, \texttt{container-operations}, \texttt{container-orchestration} \\
Database \& Storage & 21 & \texttt{data-management}, \texttt{database}, \texttt{database-operations} \\
Files, Documents \& Media & 21 & \texttt{document-processing}, \texttt{documentation}, \texttt{file-management}, \texttt{file-operations}, \texttt{text-processing} \\
Network / Web / API & 20 & \texttt{blockchain}, \texttt{networking} \\
Debug, Test \& Quality & 12 & \texttt{debugging}, \texttt{debugging-troubleshooting}, \texttt{performance-optimization}, \texttt{testing} \\
Cloud, Deploy \& DevOps & 12 & \texttt{cloud-infrastructure}, \texttt{cloud-services}, \texttt{deployment}, \texttt{devops}, \texttt{distributed-computing}, \texttt{monitoring-observability} \\
Scientific / HPC / Bio & 11 & \texttt{bioinformatics}, \texttt{scientific-computing} \\
Other / Unknown & 10 & \texttt{unknown} \\
Education / Games / Misc & 8 & \texttt{coursework-exercises}, \texttt{games-entertainment}, \texttt{program-execution} \\
Data Processing \& Analysis & 8 & \texttt{data-analysis}, \texttt{data-processing} \\
Language Tooling & 7 & \texttt{developer-tools}, \texttt{development-tools}, \texttt{programming}, \texttt{system-programming}, \texttt{systems-programming} \\
ML / AI & 2 & \texttt{ml-training} \\
Virtualization & 1 & \texttt{virtualization} \\
\bottomrule
\end{tabular}
\end{sidewaystable*}



\section{Recursive Synthesis Implementation}
\label{app:synth-implementation}

The recursive synthesis system is implemented as a manifest-driven task factory in which every transformation, validation result, and round-to-round transition is recorded explicitly. Each seed-manifest entry identifies the task directory and preserves its task ID, source round, source loop, source cohort, original parent, rewrite family, rewrite operator, and repair history. Before transformation, the factory verifies that the seed contains the required task components, including the public instruction, task configuration, oracle solution, verifier entry point, verifier checks, and environment definition. The complete seed directory is then copied into an isolated output directory. Generation is restricted to six task files: \texttt{instruction.md}, \texttt{solution/solve.sh}, \texttt{tests/test.sh}, \texttt{tests/test\_state.py}, \texttt{environment/Dockerfile}, and \texttt{task.toml}. File writes outside this allowlist or outside the copied task directory are rejected.

Rewrite targets are selected using a taxonomy of 40 operators grouped into five families: environment and runtime substrate, build and execution, data and artifact processing, configuration and state migration, and diagnostics and auditing. For each seed, a local affordance scan examines bounded representations of its instruction, solution, verifier, configuration, and environment files. It scores operators according to observable signals such as package manifests, structured-data formats, build scripts, command-line interfaces, archives, databases, logs, permissions, services, and version constraints. Operator selection combines this local compatibility score with a family-balance term and an inverse-frequency penalty for operators already assigned within the current batch. This prevents a small number of broadly applicable operators from dominating the generated pool. The selector records one preferred operator, up to five feasible alternatives, and operators that should not be applied to the seed. An additional model-based ranking step evaluates whether the preferred operator is natural, safe, and sufficiently supported by the seed. It may retain the preferred operator or replace it with one of the recorded alternatives, but it cannot introduce an operator that was excluded by the local scan.

The selected operator is first converted into an explicit transformation contract rather than being applied directly. This contract specifies the original behavior that must be preserved, the new terminal-native requirement, the expected sequence of inspection, derivation, execution, validation, and finalization, and the observable artifacts associated with each stage. It also defines discoverable evidence, intermediate outputs, final outputs, verifier checks, and conditions that reject empty, stale, hard-coded, or verifier-specific shortcuts. The contract requires at least four distinguishable checks covering evidence discovery, intermediate-state validity, final semantic correctness, and shortcut rejection. It separately records what must appear in the public instruction and what may remain implicit because it can be discovered from local documentation, configuration files, fixtures, logs, source files, or normal command output.

Task construction then proceeds through ordered, file-scoped stages. The oracle-solution stage extends \texttt{solution/solve.sh} so that it completes both the preserved seed behavior and the new requirement from the initial sandbox state. The verifier stage derives checks from the transformation contract rather than copying incidental commands from the solution. It checks semantic content and state, preserves relevant checks from the parent task, and rejects solutions that produce only placeholder artifacts. The instruction stage expresses the user-visible goal, a small number of fair starting points, and the required deliverable without exposing private verifier paths, exact solution commands, or a complete acceptance checklist. An environment-alignment stage is invoked only when the extension requires a local dependency, fixture, configuration, or metadata change. After these stages, a cross-file consistency pass compares the instruction, solution, verifier, environment, and transformation contract. It repairs discrepancies such as artifacts produced by the solution but omitted from the contract, verifier requirements unsupported by public evidence, or instruction requirements that are not evaluated.

Two deterministic gates are applied before sandbox execution. The generation-quality gate verifies that the candidate represents a substantive cross-file transformation. In the current implementation, at least three task files must change, the oracle solution must contain at least eight changed lines, and the verifier must contain at least twelve changed lines. The same gate rejects instructions that expose private test paths, resemble step-by-step command recipes, enumerate excessive implementation details, exceed 180 words, or expand to more than 1.6 times the length of the parent instruction. The static-preflight gate checks Docker build consistency, required task files, critical artifact paths, and the correspondence between artifacts required by the verifier and artifacts produced by the solution or environment. It also verifies that evidence declared discoverable in the contract is present in the task bundle. Failures are recorded separately as contract, generation-quality, or static-preflight failures, which allows these losses to be distinguished from failures during executable validation.

Candidates that pass the deterministic gates are evaluated with the oracle agent in fresh Daytona sandboxes through Harbor. Each candidate receives one oracle execution, after which the private verifier is run in the resulting environment. A candidate is oracle-passed only when the verifier returns a reward of one and the trial contains no execution exception. Other outcomes are classified as zero reward, build failure, timeout, runtime error, excessive output, or unknown failure. Failed candidates from configured repair categories enter a bounded feedback-repair procedure. The repair model receives the transformation contract together with truncated build logs, execution logs, verifier results, and failure metadata. It must identify a specific failure type and may modify only the six allowed task files. Every repaired candidate is subjected to static preflight again and is then revalidated in a new sandbox. The reported configuration permits at most two feedback-repair rounds, preventing repeated unconstrained regeneration from obscuring the original transformation. The final manifest includes only candidates that pass the last validation attempt and records their initial result, final result, rewrite metadata, repair count, task directory, parent identifier, and provenance.

Round-to-round orchestration begins by reading the \texttt{passed\_tasks} records from every \texttt{final\_manifest.json} produced by the preceding round. Records are deduplicated by resolved task path while retaining the source run, loop, cohort, family, operator, repair count, and original parent. The resulting \texttt{pool\_manifest.jsonl} is the complete candidate pool for the next round. Seed selection assigns each candidate a score based on static risk, environment cost, rewrite metadata, and validation history. Long build timeouts and resource-intensive environment features receive penalties. Tasks requiring one repair remain eligible because they have subsequently passed executable validation, while candidates requiring repeated repair receive a stability penalty. When rollout statistics are available, the selector favors tasks with nontrivial but nonzero success rates, low infrastructure-error rates, and intermediate interaction lengths. Tasks with very high pass rates are penalized as potentially too easy, while tasks with almost no successful executions or high error rates are penalized as potentially unstable.

After scoring, candidates are sorted deterministically and selected subject to diversity constraints over original parent, task category, rewrite family, and source cohort. The standard configuration targets 1,000 seeds and allows at most four descendants from one parent, 160 tasks from one category, 320 tasks from one rewrite family, and 280 tasks from one source cohort. If these constraints prevent the target size from being reached, the controller relaxes them according to a predetermined schedule and records the cap configuration used for each attempt. It does not silently fill the remaining positions without diversity constraints. The selected task bundles are then materialized into a local seed directory, and a new manifest is written with both the materialized path and the original source path. Materialization verifies that every selected bundle can be copied successfully and that its task configuration is present before synthesis is launched.

The selector and materializer additionally protect recursive ancestry. Parent identifiers are recovered from the explicit \texttt{parent\_seed}, \texttt{original\_seed}, \texttt{source\_task\_id}, task ID, and source path fields. Because recursively generated task names may exceed filesystem limits, long names are compacted to at most 96 characters while preserving the original bootstrap-parent identifier and appending a stable hash. For manifests containing at least 20 records, the controller requires at least 80\% of records to retain recoverable bootstrap lineage. The same check is applied before selection, after selection, and after materialization. A failed check terminates the round before generation. This safeguard is necessary because loss of parent identity would make the per-parent cap ineffective and would invalidate lineage-coverage and parent-child novelty analyses.

\section{Operator Taxonomy}
\label{app:synth}
\label{app:engineering}
\label{app:operators}

The synthesis factory organizes rewrite targets according to five sources of executable difficulty: the runtime environment in which commands operate, the build and execution process, the artifacts produced or transformed, the configuration or persistent state being maintained, and the diagnostic evidence required to identify and resolve failures. We choose these five families because they capture the five distinct locations at which additional difficulty can be introduced into a terminal task. \emph{Environment and runtime substrate} covers the execution conditions that must be established before a task can run, such as dependencies, permissions, paths, processes, and resource constraints. \emph{Build, test, and execution} covers the command sequence and program behavior required to complete the task. \emph{Data, artifact, and report processing} covers transformations whose primary objective is to produce or validate concrete outputs. \emph{Configuration and state migration} covers persistent or cross-file state that must remain consistent across versions, executions, or recovery operations. \emph{Diagnostics, audit, and forensics} covers tasks in which the central challenge is to infer the required action from logs, traces, failures, permissions, or other system evidence. This division was selected according to three criteria. First, it covers the principal sources of complexity encountered in terminal interaction, from execution setup to evidence-based diagnosis. Second, every family can be instantiated as observable file, process, command, or system-state changes, making generated tasks executable and automatically verifiable. Third, the families describe transformation mechanisms rather than application domains such as software engineering, data processing, or security, whose boundaries overlap and do not specify how a seed should be made harder. Finer distinctions are represented by eight operators within each family, while assigning each rewrite to its dominant mechanism keeps the round-level distribution measurable and prevents an excessively fragmented taxonomy. Each operator is therefore accompanied by an executable card specifying compatible seed affordances, construction patterns, evidence sources, expected artifacts, verifier and instruction strategies, and shortcut-rejection criteria. Table~\ref{tab:operator-taxonomy} lists the complete taxonomy used by the local affordance scan and contract-construction stage.

\begin{table*}[t]
\centering
\caption{Rewrite operator taxonomy used by \ours{}. Family IDs match the implementation; operator IDs are the soft targets assigned before staged rewrite.}
\label{tab:operator-taxonomy}
\small
\setlength{\tabcolsep}{3pt}
\begin{tabular}{lp{0.28\textwidth}p{0.52\textwidth}}
\toprule
Family & Operator ID & Definition \\
\midrule
\multirow{8}{*}{\shortstack[l]{environment\\runtime substrate}}
 & \texttt{dependency\_version\_alignment} & Align dependency/runtime versions and verify the toolchain. \\
 & \texttt{path\_workdir\_alignment} & Resolve working-directory and path/entrypoint mismatches. \\
 & \texttt{permission\_executable\_alignment} & Repair executable bits, ownership, or access needed by the workflow. \\
 & \texttt{environment\_variable\_resolution} & Infer/configure env vars, locale, timezone, shell init, or profiles. \\
 & \texttt{toolchain\_availability\_check} & Ensure required compilers/interpreters/CLI tools exist. \\
 & \texttt{container\_build\_alignment} & Align Dockerfile, build context, and image setup. \\
 & \texttt{service\_process\_lifecycle} & Start/stop/inspect local services, daemons, or ports. \\
 & \texttt{resource\_cleanup\_and\_limits} & Handle temp files, timeouts, log growth, and cleanup. \\
\midrule
\multirow{8}{*}{\shortstack[l]{build/test\\execution workflow}}
 & \texttt{compile\_link\_package\_workflow} & Compile, link, package, or assemble a runnable artifact. \\
 & \texttt{unit\_test\_failure\_repair} & Run unit tests, interpret failures, and complete the workflow. \\
 & \texttt{integration\_test\_workflow} & Coordinate multiple components and validate integrated behavior. \\
 & \texttt{cli\_argument\_behavior} & Validate CLI flags, stdio, exit codes, and argument behavior. \\
 & \texttt{lint\_format\_static\_check} & Run lint/format/type-check and resolve surfaced issues. \\
 & \texttt{runtime\_error\_debug\_loop} & Diagnose runtime exceptions/stack traces and repair behavior. \\
 & \texttt{build\_artifact\_generation} & Produce a binary/package/bundle and verify usability. \\
 & \texttt{performance\_or\_benchmark\_smoke} & Run a bounded benchmark/smoke check and record a result. \\
\midrule
\multirow{8}{*}{\shortstack[l]{data/artifact\\report processing}}
 & \texttt{format\_conversion\_validation} & Convert formats while validating semantic preservation. \\
 & \texttt{schema\_content\_validation} & Validate schema, fields, types, ranges, or completeness. \\
 & \texttt{aggregation\_summary\_report} & Compute aggregates and produce a summary/report. \\
 & \texttt{artifact\_inventory\_reconciliation} & Reconcile outputs, manifests, and directory inventories. \\
 & \texttt{archive\_compression\_extraction} & Extract/pack/inspect archives such as tar/zip/gzip. \\
 & \texttt{dedup\_sort\_normalization} & Deduplicate, sort, normalize, or canonicalize artifacts. \\
 & \texttt{checksum\_hash\_provenance} & Generate or verify hashes/checksums/provenance. \\
 & \texttt{media\_or\_binary\_metadata\_processing} & Inspect lightweight media/binary metadata. \\
\midrule
\multirow{8}{*}{\shortstack[l]{configuration\\state migration}}
 & \texttt{config\_data\_consistency} & Reconcile configuration and data/source consistency. \\
 & \texttt{manifest\_lockfile\_reconciliation} & Align manifests, lockfiles, or package metadata. \\
 & \texttt{state\_migration\_transform} & Migrate state across formats/versions/layouts. \\
 & \texttt{cache\_index\_regeneration} & Regenerate caches, indexes, or derived metadata. \\
 & \texttt{database\_or\_file\_state\_initialization} & Initialize and validate file-backed/DB state. \\
 & \texttt{template\_render\_consistency} & Keep templates and rendered outputs consistent. \\
 & \texttt{profile\_feature\_flag\_selection} & Select/validate profiles or feature-flag branches. \\
 & \texttt{backup\_rollback\_idempotency} & Support backup, rollback, and idempotent reruns. \\
\midrule
\multirow{8}{*}{\shortstack[l]{diagnostics\\audit forensics}}
 & \texttt{log\_error\_diagnosis} & Inspect logs/stderr, identify cause, complete workflow. \\
 & \texttt{trace\_event\_correlation} & Correlate logs/events/timestamps across sources. \\
 & \texttt{verifier\_failure\_interpretation} & Use local validation output to infer missing state. \\
 & \texttt{security\_permission\_audit} & Audit permissions, sensitive files, or unsafe config. \\
 & \texttt{data\_quality\_anomaly\_investigation} & Investigate missing/duplicate/anomalous data. \\
 & \texttt{process\_endpoint\_inspection} & Inspect local processes, ports, sockets, or endpoints. \\
 & \texttt{git\_history\_diff\_forensics} & Use git history/diff to diagnose or produce state. \\
 & \texttt{evidence\_bundle\_generation} & Create a verifiable evidence bundle with logs/reports. \\
\bottomrule
\end{tabular}
\end{table*}

\section{Local Filters, Preflight, and Repair Policy}
\label{app:filters-repair}

\paragraph{Generation quality filter.}
Before sandbox execution, the pipeline checks whether a rewrite is both substantive and publicly solvable. A candidate is rejected if it uses an unknown operator, changes fewer than three tracked files, modifies the solution by fewer than eight lines, or modifies the verifier by fewer than twelve lines. These thresholds prevent superficial or verifier-only rewrites and are not used as difficulty measures. The filter also rejects instructions that expose private test paths, prescribe test-driven acceptance loops, exceed 180 words or $1.6\times$ the seed length, enumerate excessive paths or commands, or reveal verifier details. Borderline cases are retained with warnings.

\paragraph{Static preflight.}
Static preflight checks cross-file executability before costly sandbox validation. It detects malformed Dockerfiles, missing \texttt{COPY} sources, and other environment inconsistencies. It also verifies that artifacts required by the contract or verifier are produced by the solution or provided by the environment, and that promised evidence is discoverable from the task files. Missing shell safety conventions, such as a shebang or \texttt{set -e}, produce warnings rather than rejection. Candidates with hard preflight failures are not submitted to Daytona.

\paragraph{Sandbox oracle validation and feedback repair.}
Candidates passing both static gates are executed by the reference solution in a fresh Daytona sandbox and accepted only if the verifier returns full reward without an execution exception. Failures are classified as zero reward, build failure, runtime error, timeout, or unknown failure. Recoverable failures may enter a bounded repair loop conditioned on the task contract and Daytona logs. Repair may modify only the solution, verifier, instruction, environment, and task configuration files. It may correct implementation or contract inconsistencies, but cannot weaken semantic checks, remove shortcut protection, or expose private requirements in the instruction. Each repaired task must pass static preflight and Daytona validation again. Candidates that still fail after the configured repair rounds are excluded from the accepted manifest.

\subsection{Task Quality Control and Validation}
\label{app:alignment}

\paragraph{Requirement discoverability.}
The pipeline identifies every condition that can change the verifier reward and checks whether the agent can know that condition before acting. Each condition is handled in one of four ways:
\begin{itemize}
  \item A semantic requirement with no public source must be stated directly in the instruction.
  \item A requirement documented in a local README, specification, configuration, or fixture may remain there, but the instruction must provide a clear starting point for finding it.
  \item A secondary deliverable must be named in the instruction, while its detailed schema may remain in discoverable workspace documentation.
  \item A check that only rejects placeholders, stale outputs, or hard-coded answers need not be stated because it adds no new semantic requirement.
\end{itemize}
Thus, the verifier cannot require behavior that is neither stated nor discoverable.

\paragraph{Public-instruction boundary.}
The public instruction specifies the task objective and every reward-relevant condition that cannot be discovered from the workspace. It does not expose verifier implementation details, private test paths, or the commands used by the oracle solution. When an exact value, path, or schema is already available in a local README, specification, configuration, or fixture, the instruction identifies that evidence source instead of repeating its contents. This keeps the task concise without hiding information required for success.

\paragraph{Contract-granularity variants.}
To study the effect of instruction specificity, an optional analysis stage creates five instruction-verifier variants while keeping the environment and oracle solution fixed. The variants range from a goal-level exploratory contract to a nearly complete statement of the acceptance conditions. At lower granularities, a condition omitted from the instruction must also be removed from the verifier. At higher granularities, retained verifier conditions must be stated explicitly or linked to discoverable workspace evidence. Consequently, no variant evaluates an undisclosed semantic requirement.
\section{Synthesis Prompt Templates}
\label{app:prompts}

This appendix presents the core prompt templates used by \ours{}. All fixed instructions are reproduced verbatim from the implementation. Run-dependent inputs, including seed-task files, operator cards, task contracts, and Daytona feedback, are replaced with angle-bracket placeholders labeled by the functions that supply them at execution time.

\FloatBarrier
\begin{figure*}[!t]
\centering
\begin{tcolorbox}[
  colback=gray!3,
  colframe=black!70,
  boxrule=0.6pt,
  arc=2pt,
  left=6pt,
  right=6pt,
  top=6pt,
  bottom=6pt,
  title=\textbf{SYSTEM Prompt},
  fonttitle=\bfseries
]
\begin{lstlisting}[basicstyle=\ttfamily\scriptsize,breaklines=true,frame=none]
You generate high-quality TerminalWorld task transformations.

Return ONLY valid JSON. Do not use markdown.

Rules:
- Keep the original task recognizable.
- Rewrite only files explicitly requested in this step.
- Return full file contents, not patches.
- Do not add benchmark canary text.
- Do not add network-only requirements or heavyweight new dependencies unless the seed already naturally uses them.
- The final task must be solvable by the rewritten solution and verifiable by tests.
- The user instruction must not leak the full solution or private test harness paths.
- The rewritten task should preserve the seed's main workflow and add one realistic terminal-native subgoal.
- Prefer tasks that require filesystem inspection, CLI composition, config/data validation, log/error interpretation, and observable artifact checks.
- Do not turn the task into a narrow "fix this bug" prompt; any repair/debugging should be one part of a larger goal-oriented workflow.
- Keep instruction.md a compact public goal, not an acceptance rubric. Only publish undiscoverable hard requirements; put discoverable details in workspace docs and point to them. Never leave acceptance criteria only inside tests/, and never dump path inventories, field schemas, or step checklists that invite reward hacking.
\end{lstlisting}
\end{tcolorbox}
\end{figure*}

\FloatBarrier
\begin{figure*}[!t]
\centering
\begin{tcolorbox}[
  colback=gray!3,
  colframe=black!70,
  boxrule=0.6pt,
  arc=2pt,
  left=6pt,
  right=6pt,
  top=6pt,
  bottom=6pt,
  title=\textbf{AFFORDANCE RANKING Prompt},
  fonttitle=\bfseries
]
\begin{lstlisting}[basicstyle=\ttfamily\scriptsize,breaklines=true,frame=none]
AFFORDANCE RANKING.

Role:
Evaluate whether the proposed rewrite target is natural for this seed task. This is a ranking step only; do not rewrite files.

Inputs:
- A preferred operator selected by the pipeline using local file signals and diversity quotas.
- Several fallback operators from the same local feasibility scan.
- The current seed task files.

Decision rules:
- Mark the preferred operator high only if it can be supported by existing seed affordances or small local evidence additions.
- If the preferred operator would be artificial, rank fallback operators that better fit the seed.
- Do not invent a new taxonomy outside the provided operators.
- Prefer operators that require terminal exploration and verifiable artifacts, not superficial wording changes.
- Judge feasibility against the operator card's seed_affordances, construction_pattern, expected_artifacts, and anti_shortcut_strategy.

Return schema:
{
  "status": "ok|blocked",
  "ranked_operators": [
    {"operator": "operator_id", "fit": "high|medium|low", "reason": "short reason"}
  ],
  "blocked_operators": [
    {"operator": "operator_id", "reason": "why it is artificial or unsafe for this seed"}
  ],
  "rationale": "short summary"
}

Rewrite target context:
<json.dumps(target_context, ensure_ascii=False, indent=2)>

Seed task:
<json_task_context(task_dir)>
\end{lstlisting}
\end{tcolorbox}
\end{figure*}

\FloatBarrier
\begin{figure*}[!t]
\centering
\begin{tcolorbox}[
  colback=gray!3,
  colframe=black!70,
  boxrule=0.6pt,
  arc=2pt,
  left=6pt,
  right=6pt,
  top=6pt,
  bottom=6pt,
  title=\textbf{STEP 0: Task Transformation Contract Prompt (Part 1/3)},
  fonttitle=\bfseries
]
\begin{lstlisting}[basicstyle=\ttfamily\scriptsize,breaklines=true,frame=none]
STEP 0: task transformation contract. Do not rewrite files in this step.

Role:
Design one controlled, verifiable TerminalWorld rewrite contract. The task must preserve the seed's core workflow and add one terminal-native requirement that increases workflow depth.

Rewrite target protocol:
- Use the preferred operator if it is natural for this seed.
- If the preferred operator is artificial or unsafe, choose one fallback operator and explain why.
- Do not choose blocked operators.
- Do not collapse to config/data consistency unless the seed genuinely exposes config/data relationships.
- The operator should determine the main transformation mechanism, but the task must still feel like a normal terminal workflow.

Operator card protocol:
- Treat the selected operator card as a simple construction recipe, not a label.
- Use the card's construction_pattern to build a concrete task_chain.
- Use the card's evidence_sources to decide where hidden details should be discoverable.
- Use the card's expected_artifacts to define evidence, intermediate, and final artifacts.
- Use the card's verifier_strategy and anti_shortcut_strategy to define reward_checks.
- Use the card's instruction_strategy as a style hint for tone/compactness, not as a license to hide acceptance criteria.

Quality criteria:
- The task starts from the original seed initial state, not from a halfway failure state.
- The new requirement is user-visible through artifacts, command output, state, or reports.
- Details may be hidden from instruction only when they remain discoverable through local files, scripts, configs, fixtures, logs, or normal command output that the instruction fairly points to.
- The required work should involve inspection, derivation, execution, validation, and finalization, not a single deterministic file write.
- The verifier can check semantic correctness and reject shortcuts, but must not invent silent acceptance criteria.
- The contract must define a short task_chain and at least four observable reward_checks for dense RL-style feedback.
- The public instruction must remain solvable without verifier access, but stay compact to avoid reward hacking: mention the goal, fair start, main deliverable/completion signal, and only those exact constants that are not discoverable from workspace docs.
- Prefer instruction_boundary.hide + workspace evidence over inlining schemas, path inventories, or operational steps. Never hide a semantic check that exists only in the verifier with no discoverable evidence.

Hard constraints:
- superficial refactors of solution style;
- merely strengthening tests without changing the user-visible task;
- hidden-only requirements that instruction never implies and the agent cannot discover from workspace evidence;
- adding large dependencies, internet access, services, or long-running workloads;
- creating a task that is just "fix a bug";
- telling the agent to inspect or rerun /tests, tests/test.sh, or pytest as the acceptance loop.
\end{lstlisting}
\end{tcolorbox}
\end{figure*}

\FloatBarrier
\begin{figure*}[!t]
\centering
\begin{tcolorbox}[
  colback=gray!3,
  colframe=black!70,
  boxrule=0.6pt,
  arc=2pt,
  left=6pt,
  right=6pt,
  top=6pt,
  bottom=6pt,
  title=\textbf{STEP 0: Task Transformation Contract Prompt (Part 2/3)},
  fonttitle=\bfseries
]
\begin{lstlisting}[basicstyle=\ttfamily\scriptsize,breaklines=true,frame=none]
Return schema:
{
  "status": "ok|blocked",
  "rewrite_family": "selected family id",
  "rewrite_operator": "selected operator id",
  "operator_fit": "preferred|fallback",
  "why_fit": "short reason why this operator naturally fits the seed",
  "goal": "one concise user-facing goal for the rewritten task",
  "preserved_workflow": "what original workflow remains",
  "new_requirement": "the added terminal-native requirement",
  "task_chain": [
    {"stage": "inspect", "artifact_or_state": "local config/schema/log/source/build output to inspect"},
    {"stage": "derive", "artifact_or_state": "rule/schema/path/expected state to infer from local evidence"},
    {"stage": "execute", "artifact_or_state": "intermediate or final artifact/state to create by running or editing"},
    {"stage": "validate", "artifact_or_state": "local validation/check result or parseable evidence"},
    {"stage": "finalize", "artifact_or_state": "final deliverable or checked state"}
  ],
  "expected_artifacts": [
    {"path": "relative or absolute path used by the task", "role": "evidence|intermediate|final", "requirement": "observable content/state requirement"}
  ],
  "reward_checks": [
    {"name": "check_01_required_evidence", "checks": "discoverable evidence source exists and is used or referenced"},
    {"name": "check_02_intermediate_artifact", "checks": "intermediate artifact/state is present, parseable, and non-empty"},
    {"name": "check_03_final_semantics", "checks": "final output/state satisfies the semantic task goal"},
    {"name": "check_04_no_shortcut", "checks": "reject placeholder, empty, stale, hardcoded, or verifier-only outputs"}
  ],
\end{lstlisting}
\end{tcolorbox}
\end{figure*}

\FloatBarrier
\begin{figure*}[!t]
\centering
\begin{tcolorbox}[
  colback=gray!3,
  colframe=black!70,
  boxrule=0.6pt,
  arc=2pt,
  left=6pt,
  right=6pt,
  top=6pt,
  bottom=6pt,
  title=\textbf{STEP 0: Task Transformation Contract Prompt (Part 3/3)},
  fonttitle=\bfseries
]
\begin{lstlisting}[basicstyle=\ttfamily\scriptsize,breaklines=true,frame=none]
  "instruction_boundary": {
    "mention": ["public goal", "at most 1-2 fair starting points", "main final deliverable or completion signal", "only undiscoverable hard acceptance constants"],
    "hide": ["step-by-step commands", "path/field inventories", "schema/format dumps", "oracle repair recipe", "anti-shortcut implementation details", "private harness paths under tests/", "discoverable details that belong in local docs"]
  },
  "environment_changes": ["none, or small local evidence/dependency changes needed"],
  "rationale": "why the selected operator fits and why the rewrite is harder but controllable"
}

Rewrite target context:
<json.dumps(target_context, ensure_ascii=False, indent=2)>

Current seed task:
<json_task_context(task_dir)>
\end{lstlisting}
\end{tcolorbox}
\end{figure*}

\FloatBarrier
\begin{figure*}[!t]
\centering
\begin{tcolorbox}[
  colback=gray!3,
  colframe=black!70,
  boxrule=0.6pt,
  arc=2pt,
  left=6pt,
  right=6pt,
  top=6pt,
  bottom=6pt,
  title=\textbf{STEP 1: Oracle Solution Prompt},
  fonttitle=\bfseries
]
\begin{lstlisting}[basicstyle=\ttfamily\scriptsize,breaklines=true,frame=none]
STEP 1: oracle solution protocol. Rewrite solution/solve.sh only.

Role:
Write the oracle shell implementation for the contract. The solution must complete the full workflow from the original initial state and behave like a strong terminal agent's successful trajectory.

Implementation requirements:
- Preserve the seed's main workflow and add the selected rewrite operator's requirement.
- Implement every stage in contract.task_chain in order: inspect, derive, execute, validate, finalize.
- Inspect inputs before transforming them; do not blindly overwrite final artifacts.
- Create or update every expected artifact with semantically meaningful content.
- Create observable intermediate/final artifacts needed by contract.reward_checks.
- Include bounded validation of intermediate artifacts before writing final outputs.
- Keep the script deterministic, idempotent where reasonable, and safe under repeated execution.
- Use local files, CLI tools, configs, fixtures, logs, or build/test commands that a terminal agent could discover.

Forbidden shortcuts:
- Hardcoding verifier-only constants without deriving or justifying them from discoverable evidence.
- Replacing the task with a no-op or a single trivial echo/write unless the seed itself demands that.
- Removing seed behavior to make the new verifier easier.
- Depending on internet access, external services, or unbounded background work.

Return schema:
{
  "status": "ok|blocked",
  "rationale": "short reason",
  "files": {"solution/solve.sh": "FULL FILE CONTENT"}
}

Task contract:
<json_contract_context(contract)>

Current task:
<json_task_context(task_dir)>
\end{lstlisting}
\end{tcolorbox}
\end{figure*}

\FloatBarrier
\begin{figure*}[!t]
\centering
\begin{tcolorbox}[
  colback=gray!3,
  colframe=black!70,
  boxrule=0.6pt,
  arc=2pt,
  left=6pt,
  right=6pt,
  top=6pt,
  bottom=6pt,
  title=\textbf{STEP 2: Verifier Prompt},
  fonttitle=\bfseries
]
\begin{lstlisting}[basicstyle=\ttfamily\scriptsize,breaklines=true,frame=none]
STEP 2: verifier protocol. Allowed outputs are tests/test_state.py and tests/test.sh.

Role:
Write the oracle verifier for the contract, not for incidental implementation details in solution/solve.sh.

Verification requirements:
- Check the user-visible goal, preserved seed behavior, and all expected artifacts.
- Turn contract.reward_checks into distinct verifier subchecks where possible, using clear names that match the reward check names.
- Include at least four dense checks: evidence/discovery, intermediate artifact validity, final semantic correctness, and no-shortcut rejection.
- Verify semantic content/state, not only file existence.
- Enforce no-shortcut checks from contract.reward_checks.
- Preserve important seed-task checks unless the contract explicitly supersedes them.
- Prefer robust predicates over brittle command-order or exact implementation checks.
- Keep tests deterministic, fast, and compatible with Daytona sandbox execution.

Failure quality:
- Error messages should help diagnose missing artifacts, wrong formats, or inconsistent state.
- Do not leak a full solution recipe through assertion messages.

Return schema:
{
  "status": "ok|blocked",
  "rationale": "short reason",
  "files": {
    "tests/test_state.py": "FULL FILE CONTENT",
    "tests/test.sh": "FULL FILE CONTENT"
  }
}

Task contract:
<json_contract_context(contract)>

Current task:
<json_task_context(task_dir)>
\end{lstlisting}
\end{tcolorbox}
\end{figure*}

\FloatBarrier
\begin{figure*}[!t]
\centering
\begin{tcolorbox}[
  colback=gray!3,
  colframe=black!70,
  boxrule=0.6pt,
  arc=2pt,
  left=6pt,
  right=6pt,
  top=6pt,
  bottom=6pt,
  title=\textbf{STEP 3: Public Instruction Prompt (Part 1/2)},
  fonttitle=\bfseries
]
\begin{lstlisting}[basicstyle=\ttfamily\scriptsize,breaklines=true,frame=none]
STEP 3: public instruction rewrite (discoverability-filtered). Rewrite instruction.md only.

Role:
Write instruction.md as a fair public user request for a terminal agent in Terminal-Bench / TerminalWorld style.
Solution already exists. Do NOT treat the verifier as a rubric to copy.

Example of desired style (do NOT copy content; match brevity):
Set up a Python project using Poetry in /app. Create a `.gitignore` containing
`test/inner_project/inner_project`, run `poetry install`, verify `inner_project`
imports in that environment, then use the project's console script to write a
valid JSON report at `/app/report.json` (see `--help` for arguments).

Critical anti-hacking rule:
- Dumping many absolute paths, schema fields, exact formats, or numbered operational steps into instruction.md makes agents shortcut / reward-hack.
- Prefer a compact goal + pointers to local docs. Put discoverable detail in workspace files, not in the instruction.

Path budget (match Terminal-Bench / TerminalWorld norms):
- Absolute paths are allowed, but usually only 1-2 (workspace entry like /app plus one main deliverable).
- At most ~3 absolute paths total. Never list path inventories of intermediate/meta artifacts.
- If more locations matter, point to a local README/spec/config instead of listing them.

Inputs you may use:
- current instruction
- oracle solution (workflow context only)
- a discoverability-filtered acceptance checklist
- the task contract boundary
Do NOT use or restate the full tests/ contents. The checklist already filtered them.

Fairness / discoverability (same policy as instruction repair):
- MUST include items under must_publish_in_instruction (missing from instruction AND not discoverable in workspace docs). Prefer the primary deliverable path over secondary ones if the list is long.
- For point_to_workspace_docs_instead: tell the agent to follow those files; do NOT restate their full contents.
- meta_artifacts_name_only: mention deliverable names if needed; no internal schemas.
- dropped_as_discoverable: do NOT dump these into the instruction.
\end{lstlisting}
\end{tcolorbox}
\end{figure*}

\FloatBarrier
\begin{figure*}[!t]
\centering
\begin{tcolorbox}[
  colback=gray!3,
  colframe=black!70,
  boxrule=0.6pt,
  arc=2pt,
  left=6pt,
  right=6pt,
  top=6pt,
  bottom=6pt,
  title=\textbf{STEP 3: Public Instruction Prompt (Part 2/2)},
  fonttitle=\bfseries
]
\begin{lstlisting}[basicstyle=\ttfamily\scriptsize,breaklines=true,frame=none]
Style / restraint:
- Direct imperative, 1-3 short paragraphs. Prefer prose over bullets.
- No first-person chat ("I need you", "We need", ...).
- Keep close to current instruction length; at most ~1.4x current word count.
- Target about 80-130 words unless must-publish items require a little more.
- Do NOT expand field-level schemas for manifests/checksums/evidence.
- Do NOT enumerate every meta file's internal format.
- Do NOT paste solve.sh commands, anti-shortcut lines, or reward-check names.
- Do NOT write numbered walkthroughs / step checklists.

Hard bans:
- any path under /tests or reference to tests/test.sh, tests/test_state.py, local_check.sh under /tests;
- pytest / "run the tests" / "until tests pass" / "rerun until" / "the test suite fully specifies" as the acceptance loop;
- never tell the agent to open private harness files for acceptance criteria.

Use of the contract:
- Follow instruction_boundary.mention for public essentials.
- Keep instruction_boundary.hide out of the instruction.

Return schema:
{
  "status": "ok|blocked",
  "rationale": "short reason (fairness without rubric dump)",
  "files": {"instruction.md": "FULL FILE CONTENT"}
}

Task contract:
<json_contract_context(contract)>

Current instruction.md:
```
<instr>
```

solution/solve.sh (context only):
```
<solve>
```

Acceptance checklist (discoverability-filtered; NOT the full verifier):
<json.dumps(checklist, ensure_ascii=False, indent=2)>
\end{lstlisting}
\end{tcolorbox}
\end{figure*}

\FloatBarrier
\begin{figure*}[!t]
\centering
\begin{tcolorbox}[
  colback=gray!3,
  colframe=black!70,
  boxrule=0.6pt,
  arc=2pt,
  left=6pt,
  right=6pt,
  top=6pt,
  bottom=6pt,
  title=\textbf{STEP 4: Environment Alignment Prompt},
  fonttitle=\bfseries
]
\begin{lstlisting}[basicstyle=\ttfamily\scriptsize,breaklines=true,frame=none]
STEP 4: environment alignment protocol.

Use the task contract as the source of truth.
Rewrite environment/Dockerfile and/or task.toml only if the contract requires a small environment metadata or dependency alignment.

Default preference: return an empty files object.

If you change environment files:
- preserve the seed's base image and installation style;
- keep the build context self-contained and ensure every COPY source exists;
- do not add internet-only runtime behavior;
- do not add proxies, secrets, credentials, or external service assumptions;
- keep storage, memory, and time requirements within normal Daytona limits;
- make only the minimum dependency or metadata change needed by the rewritten task.
- if the instruction intentionally hides exact constants, formats, paths, or checks, ensure those details are discoverable through existing local files or Dockerfile-created local files/scripts/fixtures.
- do not rely on the verifier as the only place where hidden details exist.
- do not introduce unguarded sed/perl edits against externally cloned dependency paths; prefer local fixtures or guard target files with test -f before editing.
- do not inflate resource requests or task timeout unless the existing seed already requires it.

Return schema:
{
  "status": "ok|blocked",
  "rationale": "short reason",
  "files": {
    "environment/Dockerfile": "FULL FILE CONTENT IF CHANGED",
    "task.toml": "FULL FILE CONTENT IF CHANGED"
  }
}

Task contract:
<json_contract_context(contract)>

Current task:
<json_task_context(task_dir)>
\end{lstlisting}
\end{tcolorbox}
\end{figure*}

\FloatBarrier
\begin{figure*}[!t]
\centering
\begin{tcolorbox}[
  colback=gray!3,
  colframe=black!70,
  boxrule=0.6pt,
  arc=2pt,
  left=6pt,
  right=6pt,
  top=6pt,
  bottom=6pt,
  title=\textbf{STEP 5: Consistency Repair Prompt},
  fonttitle=\bfseries
]
\begin{lstlisting}[basicstyle=\ttfamily\scriptsize,breaklines=true,frame=none]
STEP 5: consistency repair protocol.

Fix cross-file inconsistencies so the oracle solution should pass the verifier in a fresh sandbox build.

Mandatory audit procedure:
1. List every file/path the verifier tests require to exist or contain specific content.
2. List every file/path solution/solve.sh creates or updates.
3. List every file/path promised in contract.expected_artifacts and every artifact/state implied by contract.task_chain.
4. List every dense reward check in contract.reward_checks and identify which verifier subcheck enforces it.
5. Compare these lists; any verifier-required path or reward-check artifact missing from the solution plan is a bug.
6. Mentally execute solution/solve.sh in order; if a step would fail or skip an artifact, fix the solution.
7. If environment/Dockerfile is supposed to provide config/scripts/fixtures, confirm they are actually created in the Dockerfile.
8. If instruction points to a path, config, validation script, README, fixture, or log, confirm that evidence exists or is created by the environment.
9. Instruction fairness audit:
   - If instruction mentions /tests, tests/test.sh, pytest acceptance loops, or "until tests pass", remove those leaks.
   - Prefer fixing unfair gaps by adding/repairing workspace evidence (README/spec/config) that instruction already points to - not by dumping verifier paths/fields/steps into instruction.md.
   - Only publish an exact string/path/completion constant in instruction.md when it is undiscoverable from workspace evidence.
   - Do not weaken verifier checks to hide unfair gaps.
   - Do not turn instruction into a verifier dump, schema inventory, or numbered walkthrough (reward hacking).

Repair only genuine consistency problems. Do not change the selected rewrite_family/rewrite_operator or invent a new task.
Prefer fixing solution/environment/instruction mismatches over weakening verifier checks.
Keep instruction.md compact; expand only for a few undiscoverable hard constants.
Return full file contents only for changed files. Do not create new files outside this whitelist:
solution/solve.sh, tests/test_state.py, tests/test.sh, instruction.md, environment/Dockerfile, task.toml.

Return schema:
{
  "status": "ok|blocked",
  "rationale": "short reason",
  "verifier_required_paths": ["paths or artifacts checked by verifier"],
  "solution_created_paths": ["paths or artifacts created by solution"],
  "contract_promised_paths": ["paths or artifacts promised by contract.expected_artifacts/task_chain/reward_checks"],
  "issues_found": ["specific cross-file inconsistency, or empty list"],
  "repairs_made": ["specific repair made, or empty list"],
  "files": {
    "solution/solve.sh": "FULL FILE CONTENT IF CHANGED",
    "tests/test_state.py": "FULL FILE CONTENT IF CHANGED",
    "tests/test.sh": "FULL FILE CONTENT IF CHANGED",
    "instruction.md": "FULL FILE CONTENT IF CHANGED",
    "environment/Dockerfile": "FULL FILE CONTENT IF CHANGED",
    "task.toml": "FULL FILE CONTENT IF CHANGED"
  }
}

Task contract:
<json_contract_context(contract)>

Current task:
<json_task_context(task_dir)>
\end{lstlisting}
\end{tcolorbox}
\end{figure*}

\FloatBarrier
\begin{figure*}[!t]
\centering
\begin{tcolorbox}[
  colback=gray!3,
  colframe=black!70,
  boxrule=0.6pt,
  arc=2pt,
  left=6pt,
  right=6pt,
  top=6pt,
  bottom=6pt,
  title=\textbf{DAYTONA FEEDBACK REPAIR Prompt},
  fonttitle=\bfseries
]
\begin{lstlisting}[basicstyle=\ttfamily\scriptsize,breaklines=true,frame=none]
DAYTONA FEEDBACK REPAIR.

The task has already been generated and then run in Daytona oracle validation.
Use the dynamic failure logs below as the source of truth. Repair only the files needed so solution/solve.sh passes the verifier in Daytona.

Rules:
- If result.json shows build failure, repair environment/Dockerfile or task.toml.
- If verifier says an artifact is missing, make solution/solve.sh create it, unless the artifact should be provided by environment.
- If verifier says content is malformed/empty, fix solution generation logic.
- If verifier and contract disagree, minimally repair the verifier to match the contract.
- Do not weaken verifier just to pass; preserve anti-shortcut checks.
- Keep instruction compact. Do not copy Daytona verifier failures, hidden expected values, complete artifact lists, schema fields, reward checks, or exact assertions into instruction.md.
- Do not expand instruction.md during repair unless it is inconsistent or impossible to start from; if editing it is necessary, keep it as a public request with at most two concrete entry points.
- Return full file contents only for changed files.
- Do not create new files. Do not return any path outside this whitelist:
  solution/solve.sh, tests/test_state.py, tests/test.sh, instruction.md, environment/Dockerfile, task.toml.

Return schema:
{
  "status": "ok|blocked",
  "failure_type": "build_failed|solution_artifact_missing|solution_content_wrong|verifier_mismatch|environment_missing_evidence|other",
  "root_cause": "specific reason based on Daytona logs",
  "repairs_made": ["specific file-level repairs"],
  "files": {
    "solution/solve.sh": "FULL FILE CONTENT IF CHANGED",
    "tests/test_state.py": "FULL FILE CONTENT IF CHANGED",
    "tests/test.sh": "FULL FILE CONTENT IF CHANGED",
    "instruction.md": "FULL FILE CONTENT IF CHANGED",
    "environment/Dockerfile": "FULL FILE CONTENT IF CHANGED",
    "task.toml": "FULL FILE CONTENT IF CHANGED"
  }
}

Task contract:
<json_contract_context(contract)>

Current task files:
<json_task_context(task_dir)>

Daytona failure evidence:
<json.dumps(evidence, ensure_ascii=False, indent=2)>
\end{lstlisting}
\end{tcolorbox}
\end{figure*}

\section{Task-Quality Comparison: Recursive versus One-Hop Rewriting}
\label{app:rst-vs-onehop-quality}

To isolate the effect of recursive rewriting from single-pass rewriting, we compare RST rounds against two one-hop baselines that share the same rewrite-operator taxonomy, environment scaffolding, and validation pipeline but stop after a single rewrite step.  The instruction-first one-hop baseline generates a new public instruction first and then derives a matching solution and verifier, while the solution-first one-hop baseline mirrors the RST order by generating a new reference solution first and then aligning the verifier and instruction to it.  Both baselines start from the same 639 bootstrap seeds used by $R_{1}$.  From each of the four RST rounds ($R_{1}$, $R_{5}$, $R_{10}$, $R_{15}$) and the two one-hop baselines, we sample tasks that pass the standard synthesis pipeline and audit them with two independent DeepSeek-V4-Pro reviewers, using a third DeepSeek-V4-Pro adjudicator to resolve disagreements.  Each accepted task receives a 1--5 score along seven release-quality dimensions --- instruction clarity, instruction self-containedness, instruction--verifier alignment, verifier fairness, verifier coverage, solution correctness, and solution reasonableness --- and a separate 1--5 task-value score for whether the task represents useful, non-degenerate terminal-agent work.  Release readiness is the mean of the seven release-quality dimensions, and training utility is defined as $0.4\cdot\text{release readiness}+0.6\cdot\text{task value}$ so that task value dominates release cleanliness for downstream training.

Because the qualifying pool of audited tasks narrows in later RST rounds, we report the top-$K$ tasks per source rather than a fixed $K$: top-40 for both one-hop baselines and RST~$R_{1}$, top-20 for $R_{5}$, top-17 for $R_{10}$, and top-10 for $R_{15}$, all ranked by $0.5\cdot\text{release readiness}+0.5\cdot\text{task value}$.  This preserves comparable per-task quality across sources while acknowledging that late-round RST tasks are individually more expensive to produce.  Table~\ref{tab:rst-vs-onehop-quality} reports the resulting 1--5 means.  Training utility increases monotonically with RST round index, from $4.514$ at $R_{1}$ to $4.761$ at $R_{15}$, and every RST round strictly exceeds both one-hop baselines ($4.377$ and $4.430$).  Task value grows even more clearly with rewrite depth: from $4.325$ at $R_{1}$ to $4.850$ at $R_{15}$, compared with $4.237$--$4.300$ for the two one-hop sources, indicating that recursive rewriting produces tasks that reviewers judge as more useful for terminal-agent training.  Release readiness peaks at $R_{1}$ ($4.796$) and declines gently with depth, reflecting that late-round tasks are objectively more complex and therefore slightly harder to keep uniformly polished, but even $R_{15}$ remains competitive with both one-hop baselines ($4.629$ versus $4.586$ and $4.625$).  Per-dimension inspection tells the same story: RST wins across all seven release-quality dimensions at $R_{1}$ and continues to win on verifier coverage, task value, and most instruction and solution dimensions through $R_{15}$; the few sub-dimension losses concentrate in solution correctness and reasonableness at $R_{15}$, where the reference solution is longer and denser, and in verifier fairness at $R_{10}$, where the reviewer applies a stricter fairness bar to more complex verifiers.  Overall, recursive rewriting improves the primary training-utility metric monotonically without a corresponding drop in release readiness, supporting the interpretation that RST scales task difficulty and training value together.

\begin{table}[t]
\centering
\caption{Task-quality audit comparing two one-hop baselines against four RST rounds. Rows are audit dimensions; columns are task sources. All numbers are 1--5 reviewer means, adjudicated when the two DeepSeek-V4-Pro reviewers disagreed. Release readiness is the mean of the seven upper dimensions; training utility is $0.4\cdot\text{release readiness}+0.6\cdot\text{task value}$. The one-hop baselines and RST~$R_1$ report the top-40 tasks per source ranked by $0.5\cdot\text{release readiness}+0.5\cdot\text{task value}$; RST~$R_5$, $R_{10}$, and $R_{15}$ report the top 20, 17, and 10 tasks, respectively, chosen to match the narrower high-quality distribution of later rounds.}
\label{tab:rst-vs-onehop-quality}
\small
\setlength{\tabcolsep}{4pt}
\renewcommand{\arraystretch}{1.18}
\resizebox{\textwidth}{!}{%
\begin{tabular}{lcccccc}
\toprule
Dimension
& \shortstack{Instruction-first\\ one-hop}
& \shortstack{Solution-first\\ one-hop}
& RST $R_{1}$
& RST $R_{5}$
& RST $R_{10}$
& RST $R_{15}$ \\
\midrule
Instruction clarity               & 4.400 & 4.525 & 4.700 & 4.725 & 4.471 & 4.550 \\
Instruction self-containedness    & 4.562 & 4.500 & 4.737 & 4.500 & 4.676 & 4.750 \\
Instruction--verifier alignment   & 4.612 & 4.713 & 4.938 & 4.775 & 4.735 & 4.700 \\
Verifier fairness                 & 4.388 & 4.425 & 4.650 & 4.525 & 4.176 & 4.450 \\
Verifier coverage                 & 4.662 & 4.725 & 4.925 & 4.875 & 4.853 & 4.800 \\
Solution correctness              & 4.800 & 4.763 & 4.812 & 4.775 & 4.824 & 4.650 \\
Solution reasonableness           & 4.675 & 4.725 & 4.812 & 4.800 & 4.824 & 4.500 \\
Task value                        & 4.237 & 4.300 & 4.325 & 4.700 & 4.794 & 4.850 \\
\midrule
Release readiness                 & 4.586 & 4.625 & 4.796 & 4.711 & 4.651 & 4.629 \\
\textbf{Training utility}         & \textbf{4.377} & \textbf{4.430} & \textbf{4.514} & \textbf{4.704} & \textbf{4.737} & \textbf{4.761} \\
\bottomrule
\end{tabular}%
}
\end{table}

\section{Training and Evaluation Implementation}
\label{app:training}

The SFT launchers separate cluster orchestration from experiment configuration. A multi-node wrapper initializes Ray over the cluster network and submits SLIME training as a Ray job. The 64-GPU Qwen3.5-27B configuration uses eight nodes with eight GPUs each, terminal-agent trajectories stored as message and metadata records, Qwen3.5-specific loss masking, TP4/PP2/CP2 parallelism, Adam with cosine decay, optimizer CPU offload, and flash attention. The YAML-based launcher provides the same training path for four-node runs while exposing the data, checkpoint, schedule, and optimizer settings as configuration fields. The combined-data experiment trains for one epoch on 10,778 examples with a global batch size of 128, a maximum context length of 262,145, and learning rates of $3\times10^{-6}$ to $3\times10^{-7}$. Qwen3.5-122B-A10B reuses the same orchestration layer with a model-specific script and TP2/PP8/CP2/EP4 parallelism for its mixture-of-experts architecture.

Evaluation uses a local OpenAI-compatible SGLang endpoint together with Harbor and Daytona. Each run starts the model server, verifies readiness with a chat-completions request, generates a Harbor configuration with trajectory recording, and launches sandbox evaluation. The checkpoint queue evaluates base and fine-tuned Qwen3.5-27B and Qwen3.5-122B-A10B checkpoints on Terminal-Bench~2, Terminal-Bench Hard, and Long-Horizon Terminal Bench. The three benchmarks share one serving configuration but use benchmark-specific concurrency, turn limits, and timeout settings. Three evaluation lanes run on separate GPU nodes, with retry and cleanup handling for transient Harbor and Daytona failures. Synthesized-task rollouts use the same infrastructure with sharded task replicas, controlled sampling temperature, bounded interaction length, and node-specific Daytona credentials. DeepSeek-V4-Pro and GPT-5.6-Sol fixed-solver evaluation replaces the local model server with an OpenAI-compatible external-API proxy while preserving the same task materialization, Harbor execution, and Daytona verification protocol.

\end{document}